\documentclass{article}

\usepackage{microtype}
\usepackage{graphicx}
\usepackage{subcaption}
\usepackage{booktabs} 

\usepackage{hyperref}

\usepackage[preprint]{icml2026}

\usepackage{amsmath}
\usepackage{amssymb}
\usepackage{mathtools}
\usepackage{amsthm}

\usepackage[capitalize,noabbrev]{cleveref}

\theoremstyle{plain}

\theoremstyle{definition}

\theoremstyle{remark}

\usepackage[textsize=tiny]{todonotes}

\icmltitlerunning{Dual-Granularity Contrastive Reward via Generated Episodic Guidance for Efficient Embodied RL}

\usepackage{tcolorbox}
\usepackage{xcolor}

\definecolor{promptgray}{RGB}{245,245,245}   
\definecolor{titlegray}{RGB}{90,90,90}        
\usepackage{array}

\begin{document}

\twocolumn[
  \icmltitle{Dual-Granularity Contrastive Reward via Generated Episodic Guidance for Efficient Embodied RL
  \\}
    


  \icmlsetsymbol{equal}{*}

  \begin{icmlauthorlist}
    \icmlauthor{Xin Liu}{yyy,comp}
    \icmlauthor{Yixuan Li}{yyy,comp}
    \icmlauthor{Yuhui Chen}{yyy,comp}
    \icmlauthor{Yuxing Qin}{yyy,comp}
    \icmlauthor{Haoran Li}{yyy,comp}
    \icmlauthor{Dongbin Zhao}{yyy,comp}
    
  \end{icmlauthorlist}

  \icmlaffiliation{yyy}{State Key Laboratory of Multimodal Artificial Intelligence Systems,
  Institute of Automation, Chinese Academy of Sciences}
  \icmlaffiliation{comp}{School of Artificial Intelligence, University of Chinese Academy of Sciences}

  \icmlcorrespondingauthor{Haoran Li}{lihaoran2015@ia.ac.cn}

  \icmlkeywords{Machine Learning, ICML}

  \vskip 0.3in
]



\printAffiliationsAndNotice{}  

\begin{abstract}

Designing suitable rewards poses a significant challenge in reinforcement learning (RL), especially for embodied manipulation. 
Trajectory success rewards are suitable for human judges or model fitting, but the sparsity severely limits RL sample efficiency. While recent methods have effectively improved RL via dense rewards, they rely heavily on high-quality human-annotated data or abundant expert supervision. To tackle these issues, this paper proposes Dual-granularity contrastive reward via generated Episodic Guidance (DEG), a novel framework to seek sample-efficient dense rewards without requiring human annotations or extensive supervision. Leveraging the prior knowledge of large video generation models, DEG only needs a small number of expert videos for domain adaptation to generate dedicated task guidance for each RL episode. 
Then, the proposed dual-granularity reward that balances coarse-grained exploration and fine-grained matching, will guide the agent to efficiently approximate the generated guidance video sequentially in the contrastive self-supervised latent space, and finally complete the target task. Extensive experiments on 18 diverse tasks across both simulation and real-world settings show that DEG can not only serve as an efficient exploration stimulus to help the agent quickly discover sparse success rewards, but also guide effective RL and stable policy convergence independently.

\end{abstract}

\section{Introduction}

Although supervised imitation learning (behavior cloning) has shown considerable policy-shaping capabilities in robotics control \cite{diffusionpolicy,diffusion-bc}, autonomous driving \cite{driving,driving2}, and the currently booming field of large language model fine-tuning \cite{lora,selu}, two factors severely limit its upper bound of policy performance and application scenarios: (i) its reliance on large amounts of expert data, and (ii) its susceptibility to compounding errors \cite{il-survey}.
In contrast, RL can effectively address these two issues, making it a focus once again after game artificial intelligence (AI) \cite{atari-rl}. Despite its advantages, RL also faces a tricky problem: it requires sufficient environmental interactions to support the agent’s trial-and-error and policy refinement.
To solve this problem, researchers have made numerous attempts in different directions, including improving representation modules \cite{drq-v2,curl}, expanding training data \cite{mwm,dreamerpro,dreamerv3}, exploring multi-task potential \cite{jingbo,comsd}, and so on. These advances demonstrate that, driven by well-designed rewards, RL is able to support sample-efficient policy learning processes in various scenarios.

\begin{figure}[t]
    \centering
    \includegraphics[width=0.45\textwidth]{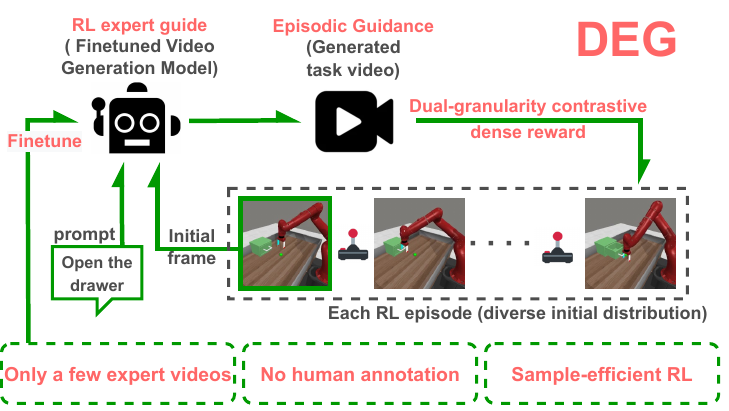}
    
    \caption{The pipeline of the proposed DEG. Without requiring human annotations or extensive supervision, DEG enables sample-efficient RL via dual-granularity contrastive dense reward based on the generated episodic video guidance. }
    \label{pipeline}
    \vspace{-3mm}

\end{figure}

However, how to design appropriate rewards for RL poses another challenge. The most primitive approach is to assign binary success-or-failure rewards to entire trajectories \cite{hilserl}. Such rewards are not only easy for humans to judge or provide feedback on, but also simple for (large) reward models to learn and fit \cite{roboreward}. Yet their sparsity often leads to low sample efficiency in RL, especially in domains with large state and action spaces.
Therefore, suitable dense rewards are often the key to achieving sample-efficient RL. Currently, common public benchmarks \cite{dmc,atari-rl} in the RL field usually provide effective dense feedback, but this is often the result of painstaking manual design by human experts \cite{procgen}. In manipulation tasks with stage-varying goals and complex interactions \cite{metaworld}, manually designing dense rewards is even quite challenging for experts \cite{metaworld-v3}.
Thus, some recent progress has aimed to realize automated dense reward design for manipulation tasks, attempting to derive dense rewards from expert demonstrations by leveraging advanced techniques such as adversarial learning \cite{GAIFO-like-ICRA22}, generative AI \cite{tevir}, and vision-language large models \cite{robodopamine}. Nevertheless, these methods either rely on large volumes of expert supervision, or require massive amounts of manually annotated data.

The above discussion naturally leads to a question: \textit{Can we design dense rewards for sample-efficient RL with minimal use of expert data and no reliance on manual annotations?} A novel framework named Dual-granularity contrastive reward via generated Episodic Guidance (DEG) is proposed to answer this open question. 
DEG leverages the prior knowledge of large models to overcome the challenge of scarce data. It finetunes open-source video generation models with a tiny number of expert videos (only 3-5 clips) for domain adaptation, familiarizing the generator with the dynamics and workflow of target tasks. 
The fine-tuned video generator then acts as a guide, producing an expert guidance video tailored to each episode with diverse initial states in the RL process. 
Via contrastive learning, DEG aligns the noisy generated videos with high-fidelity real observations, and maps the semantic distance between observations to their self-supervised latent distance. In the latent space, DEG calculates a novel dual-granularity reward that comprises a coarse-grained exploration incentive and a fine-grained matching incentive.
The former drives the agent to roughly imitate the guidance video sequentially, which is achieved via a relatively loose threshold judging imitation progress. The latter, by contrast, focuses on rewarding precise imitations that closely match the guidance video within the rough movements. 
It ensures the attention on fine-grained details (such as delicate interactions with objects), leading to the refinement of the policy from rough to accurate.
Through the organic integration of these two components, the dual-granularity reward guides the agent to efficiently and accurately approximate the generated guidance and ultimately complete the target task, as shown in Figure \ref{pipeline}.

We conduct extensive experiments on 18 manipulation tasks across both simulation and real-world settings. DEG can serve as an efficient intrinsic stimulus to help the agent quickly discover sparse success rewards, matching (or even outperforming) manually designed dense rewards by human experts. It can also guide effective RL and achieve stable policy convergence independently, surpassing state-of-the-art reward engineering approaches with much less supervision required. We summarize the contributions of our paper as follows:

\begin{itemize}
    \item We propose a novel framework of dense reward design, dubbed Dual-granularity contrastive reward via generated Episodic Guidance (DEG), for efficient embodied RL. With minimal use of expert videos and no reliance on manual annotations, DEG reward can not only help the agent quickly discover sparse success rewards, but also independently guide effective RL and achieve stable policy convergence.

    \item Through finetuning open-source video generation models, DEG requires only a few expert videos to generate dedicated task guidance for each RL episode with diverse initial states. A novel dual-granularity reward that balances coarse-grained exploration and fine-grained matching, will drive the agent to efficiently imitate the guidance in the contrastive latent space and finally complete the target task.
    
    \item We conduct extensive experiments on a set of challenging manipulation tasks, in both simulation and real-world settings. With the aid of sparse success feedback, DEG enables efficient RL across all 18 tasks, matching the performance of dense rewards handcrafted by human experts. In addition, DEG also outperforms state-of-the-art baselines on 12 reward-free tasks, with much less expert supervision required.
    

\end{itemize}

\section{Related Works}

\subsection{Manipulation Policy Learning}
Robotic manipulation has seen significant progress through behavior cloning, with recent architectures like Diffusion Policy \cite{diffusionpolicy} and ACT \cite{robo2} demonstrating high-performance imitation in modeling complex behavior distributions from expert data. However, BC remains constrained by the quality of demonstrations and struggles with sub-optimal data \cite{ppo,robo7}. To enable self-improvement, RL is often employed for policy fine-tuning \cite{ppo,hilserl,robo7}. By interacting with the environment and receiving reward signals, agents can refine their behavior beyond the expert demonstrations. Despite its potential, RL manipulation tasks are frequently hindered by the "reward engineering" bottleneck, where defining reward signals in unstructured settings requires manual effort or privileged state information that is not practical \cite{conrft,robo6}. To overcome this, our approach focuses on synthesizing reward signals for manipulation policy improvement.

\subsection{Learning from Videos}

Based on the relationship with the task domain, videos can be divided into cross-domain videos and in-domain videos. Cross-domain videos are usually abundant in quantity, but there is a gap between them and the target task in terms of observation and action spaces \cite{mvp,crptpro}. Therefore, some existing works often use them as auxiliaries to improve the performance or efficiency of existing policy learning pipelines, such as enhancing cross-domain representation \cite{mvp} or pre-training universal latent representations \cite{lapa,anotherlapa}.
In contrast, in-domain videos have an extremely high density of available information, but they are difficult to obtain directly from the Internet and need to be manually acquired at a high cost. Although many works have proved that expert videos can serve as the main \cite{lapo,FICC} or even the only source of supervision signals \cite{laifo,bcv-lr,tevir} to directly support the policy convergence on target tasks, they often rely on the assumption of sufficient video data and ignore the key metric of video data efficiency \cite{patchail,upesv}. With the rapid development of large models, it has become possible to perform large-scale pre-training on cross-domain videos to generate in-domain videos \cite{wan,seaweed,hunyuan}. Recent progress \cite{dreamgen} clearly demonstrates the feasibility of using video generation models as data generators to enhance video behavior cloning. Inspired by this, DEG first proposes to use a pre-trained video generator as an episodic guide and provides corresponding rewards during the RL process. 

\subsection{RL Reward Design}
Exploration is crucial in RL, which motivates researchers to investigate intrinsic rewards for enhancing exploration \cite{rnd}. These rewards are generally decoupled from specific tasks and proportional to the agent's exploration of the state space \cite{apt}. They can serve as plug-and-play modules to boost early-stage RL exploration \cite{protorl,crptpro} or collect unsupervised pre-training data for multi-task RL \cite{jingbo}. However, the task-agnostic nature results in their limited ability to handle complex state spaces and solve complex long-horizon tasks \cite{tevir}, which has spurred numerous works to focus on task-specific reward design. Inverse RL methods \cite{GAIFO,C-LAIFO} automatically infer expert rewards from expert demonstrations. They enable high-quality imitation of expert policies without additional human annotations while requiring abundant expert data and extensive environmental interactions \cite{il-survey}. Large reward models have achieved remarkable performance recently in both sparse feedback \cite{roboreward,roboclip,additional1} and dense incentive design \cite{robodopamine,additional2}, but they either suffer from reward hacking or rely on massive human annotations and training resources. In contrast, DEG leverages video generation models with prior knowledge as RL guides to realize automatic and accurate reward design, while breaking free from the reliance on large-scale expert data and human annotations.

\section{Methodology}

The goal of this paper is to derive the task-specific dense reward function for an environment with only success sparse reward or even no reward. The former environment can be represented as a sparse-reward Markov Decision Process (MDP) \cite{mdp} $\mathcal{M}= (\mathcal{O},\mathcal{A},\mathcal{P},\mathcal{R}^{s}, \gamma, d_0)$, where $\mathcal{O}$ denotes the visual observation space, $\mathcal{A}$ denotes the action space, $\mathcal{P}$ denotes the transition function, $\mathcal{R}^{s}$ denotes the sparse success reward function, $\gamma$ is the discount, and $d_0$ is the distribution of the initial observation. The latter can be correspondingly expressed as a reward-free MDP $\mathcal{M}= (\mathcal{O},\mathcal{A},\mathcal{P}, \gamma, d_0)$  where the symbols share the same meanings as those in the former. With the help of a well-pretrained video generation model $G$, DEG only requires a few expert videos $\mathcal{V}$ for reward calculation. 

DEG first obtains a robust task-specific guide $G'$ based on a well pre-trained video generation model $G$ and a small number of expert videos $\mathcal{V}$ (\textbf{Section 3.1}). The guide $G'$ will generate a unique task video for each episode with diverse initial states in the subsequent RL process. Then, DEG adopts contrastive self-supervised learning (\textbf{Section 3.2}) to (i) alleviate the impact of generated noise and (ii) attempt to align the semantic distance and latent distance between observations. In the self-supervised latent space, DEG calculates a novel dual-granularity dense reward (\textbf{Section 3.3}). This reward balances coarse-grained exploration and fine-grained matching, driving the agent to efficiently and accurately follow the generated instruction and finally complete the target task.

\subsection{Training Task Guide}
\begin{figure*}[t]
    \centering
    \includegraphics[width=0.95\textwidth]{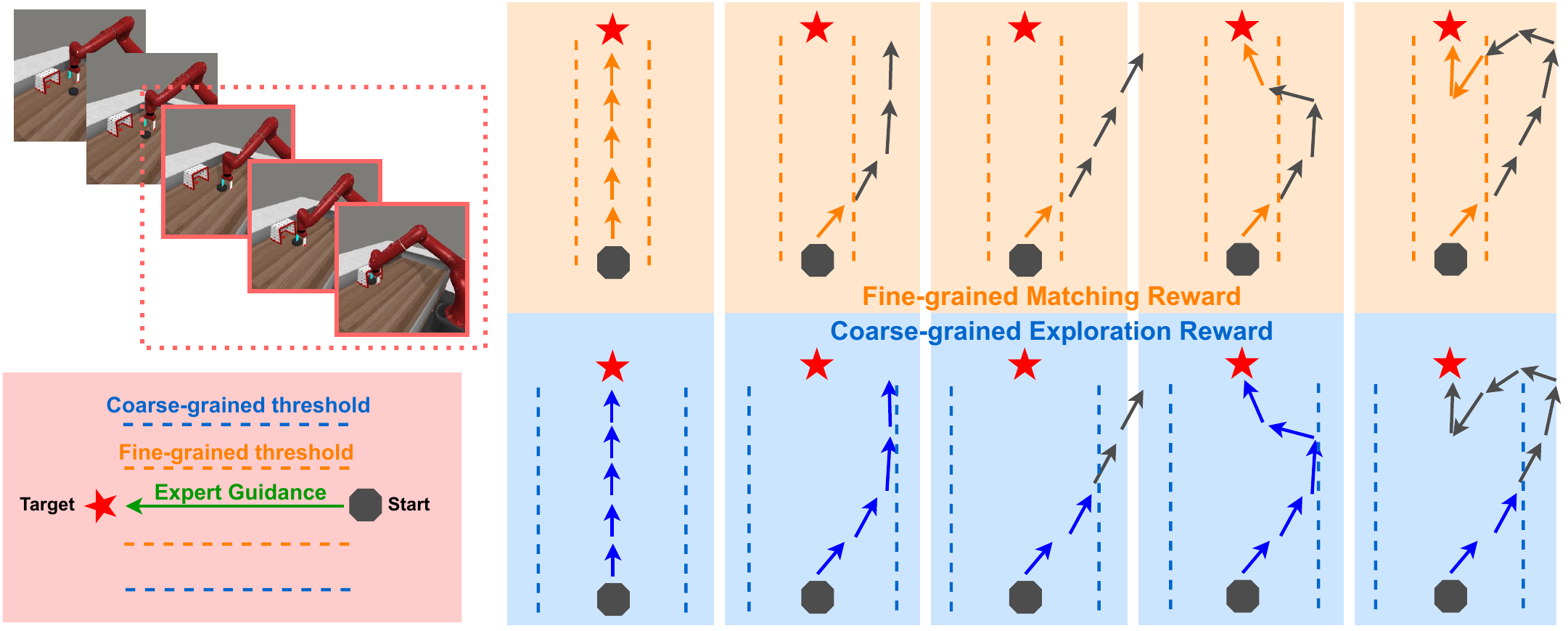}
    
    \caption{The effect diagram of coarse-grained exploration reward and fine-grained matching reward. The top-left panel shows a schematic of the expert trajectory for the \textit{plate slide} task. We take the 2D trajectory of the task's latter half (the arm pushing the plate to the target on the plane) as an example (bottom-left), with the expert guidance, coarse-grained threshold, and fine-grained threshold represented in different colors. Note that this simple example is set to clearly illustrate the core idea of our method. In practical tasks, the expert trajectories are complex 3D curves. The right panel illustrates how different trajectories trigger the two rewards, respectively. Coarse-grained rewards use a larger threshold to guide sequential target imitation, which encourages the robotic arm to roughly mimic the movement intent in the guidance. However, (i) the larger threshold tolerates operations with deviations, making it hard to learn precise interactions ('reaching target' in this task); (ii) its sequential imitation goals mean that trajectories that deviate at first but later correct and even succeed will no longer receive rewards. Fine-grained rewards directly tackle these two problems: they not only prioritize rewarding precise interactions to further refine the policy, but also ensure that trajectories achieving final success without sequential imitation receive positive feedback, thus reducing agent confusion.}
    \label{dual-reward}
   
\end{figure*}

We leverage the extensive prior knowledge of pre-trained large models to mitigate the problem of incomplete expert data distribution. Specifically, we select Wan2.1-I2V-14B \cite{wan}, an advanced and open-source image-to-video model, as our backbone. We find that without finetuning, it already possesses the ability to understand the semantics of environments and objects, while lacking knowledge on the specific dynamics relevant to the target task (e.g., the degrees of freedom and movement modes of the arm). To this end, we finetune the backbone model with the given expert videos via Low-Rank Adaptation (LoRA) \cite{lora}, adapting it to the target domain:

\begin{equation}
G' = \text{LoRA}(G; \mathcal{V}, prompt),
\end{equation}

\noindent where the prompt corresponds to the specific task. The finetuned video generator $G'$ serves as an expert guide in the RL process, generating personalized guidance $V’$ tailored to each RL episode:

\begin{equation}
 V’ = G'(o_0;prompt),
\end{equation}

where $o_0 \sim d_0$ denotes the initial observation of the current episode.  $V’$ consists of several generated frames $\{{v'}_i\}^l_{i=0}$, where $l$ denotes the fixed length of generated videos.

\subsection{Contrastive Latent Alignment}

After obtaining the expert guide, DEG aims to drive the agent to match the generated expert video sequentially to complete the target task, which can be achieved by rewarding the distance variation between the current observation and the generated frames.
Considering that the distance calculated directly in the raw pixel space cannot align with the actual semantic distance \cite{apt,protorl}, a natural approach is to leverage open-source pre-trained encoders trained on large-scale datasets \cite{dinov3} and compute the distance in the semantic latent space. However, the generated videos inevitably contain noise, and we find that pre-trained models are highly sensitive to such noise, failing to align noisy generated images (e.g., blurry or distorted ones) with their semantically identical real observations. In addition, the pre-trained encoder fails to encode the subtle observational variations in the target domain—which are negligible compared to the vast prior knowledge—into distinct changes in the latent space distance, leading to potential reward misalignment. For these issues, DEG employs self-supervised learning to train an encoder $E$ tailored to the target domain, seeking to ignore the non-semantic gaps and map the semantic distance to the computable latent distance for reward design.

Concretely, DEG employs contrastive learning \cite{atc,curl} to align expert video frames that are manually added with noise yet semantically consistent and push away different frames. A batch of images $\{{v}_i\}^M_{i=1}$ is sampled from the expert video set $\mathcal{V}$. Each ${v}_i$ is randomly augmented with noise (random shift in this paper) to obtain two noisy images. 
They are separately encoded by $E$ and its momentum copy $E'$ (updated by Exponential Moving Average (EMA) \cite{ema}) to obtain two latent features, $\hat{z}_i$ and $\check{z}_{i}$. Note that $E$ consists of both a Convolutional Neural Network (CNN) for image understanding and a Multi-Layer Perceptron (MLP) for self-supervised projection. The formula for contrastive learning is given as follows:

\begin{equation}
    \mathcal{L}_{align} = - \log \frac{\exp (u(\hat{z}_i)^\top W {\check{z}_{i}}')}{\sum_{j=1}^M\exp (u(\hat{z}_i)^\top W {\check{z}_{j}}')},
\end{equation}

where $u(\cdot)$ is another MLP set to introduce asymmetry for avoiding collapse to trivial solutions and $W$ is the contrastive matrix that is co-trained with $E$ and $u(\cdot)$.

\subsection{Dual-Granularity Dense Reward}

With the encoder $E$ trained via contrastive self-supervised learning, both generated guidance and real observations are encoded into the aligned self-supervised latent space where DEG achieves reward calculation. Concretely, DEG takes the cosine similarity between the agent observations and the generated videos $\{{v'}_i\}^l_{i=0}$ in the latent space as the metric and reward design basis:

\begin{equation}
    S(t,i) = \text{cos\_sim}(E(v'_i), E(o_t)) = \frac{E(v'_i)^\top E(o_t)}{\|E(v'_i)\| \cdot \|E(o_t)\|},
\end{equation}

where $S(t,i)$ denotes the cosine similarity between observation $o_t$ at time $t$ and the generated frame $v'_i$ indexed $i$.

The proposed dual-granularity reward comprises two components: a coarse-grained exploration reward and a fine-grained matching reward. They are both based on the distance function $S(t,i)$ while achieving distinct effects through different distance thresholds. 
The coarse-grained exploration reward uses a loose judgment threshold to encourage the agent to roughly and sequentially imitate the robotic motion trajectory in the expert guidance. This not only prevents the agent from being confined to a single trajectory (there are obviously multiple feasible trajectories for the robotic arm to reach the target object, but the expert guidance video only provides one) but also significantly reduces the learning difficulty and thus improves sample efficiency.
However, relying solely on the coarse-grained reward cannot support the agent to perform fine-grained operations in tasks requiring precise interaction with objects, which motivates us to further design the fine-grained matching reward. With a strict judgment threshold, this reward incentivizes agent observations that sufficiently match any guidance frame that is not reached yet, thus fully rewarding the accurate movements or successful interactions with objects. 
Through the organic integration of the two components, the dual-granularity reward achieves a balance between sampling efficiency and imitation accuracy. See Figure \ref{dual-reward} for a concrete example to grasp our motivation on these two rewards.

\paragraph{Coarse-grained exploration reward} drives the agent to roughly imitate the generated guidance $\{{v'}_i\}^l_{i=0}$ in sequence, thereby enabling efficient learning of the general motion intent of the robotic arm and achieving task-relevant intrinsic exploration. In each episode, we maintain a variable $I_{\text{target}}$ to represent the index of the current target frame that the agent needs to approach. This variable is initialized as
$I_{\text{target}} \leftarrow 1$ and our coarse-grained exploration reward $\mathcal{R}_{coarse}$ is formulated as the similarity gain after reaching the current observation:

\begin{equation}
    \mathcal{R}_{coarse} = S(t,I_{\text{target}})-S(t-1,I_{\text{target}}).
\end{equation}

A coarse-grained threshold $\tau_{coarse}$ is set to determine whether the target index $I_{\text{target}}$ can be sequentially advanced to the next target by fixed step number $s$:

\begin{equation}
I_{\text{target}} \leftarrow
\begin{cases}
\min(I_{\text{target}} + s,l), & \text{if } S(t,I_{\text{target}}) > \tau_{\text{coarse}} \\
I_{\text{target}}, & \text{otherwise},
\end{cases}
\end{equation}

\paragraph{Fine-grained matching reward} encourages observations that sufficiently match the expert video, thus rewarding the robotic arm for accurate movements or successful interactions with objects. Considering that (i) the trajectories for reaching the target object can be highly diverse and different with guidance and (ii) the poses required for object interaction are rather consistent and semantically aligned with the corresponding frames in the generated guidance, we don't adopt a strictly sequential target update mechanism, which is employed in coarse-grained reward. Instead, we calculate the cosine similarity between the current observation $o_t$ and the whole guidance $\{{v'}_i\}^l_{i=0}$, identifying the generated frame that is most similar to the current observation:

\begin{equation}
i^* = \arg\max_{i \in \{1,2,\dots,l\}} S(t, i).
\end{equation}

Then, a strict fine-grained threshold $\tau_{\text{fine}}$ is employed to determine whether $o_t$ sufficiently matches ${v'}_{i^*}$, and then grants a reward accordingly:

\begin{equation}
    R_{fine} =
\begin{cases}
i^*, & \text{if } S(t, i^*) > \tau_{\text{fine}} \quad \& \quad i^* > I_{\text{reached}} \\
0, & \text{otherwise},
\end{cases}
\end{equation}

\begin{figure*}[t]
    \centering
    \includegraphics[width=0.91\textwidth]{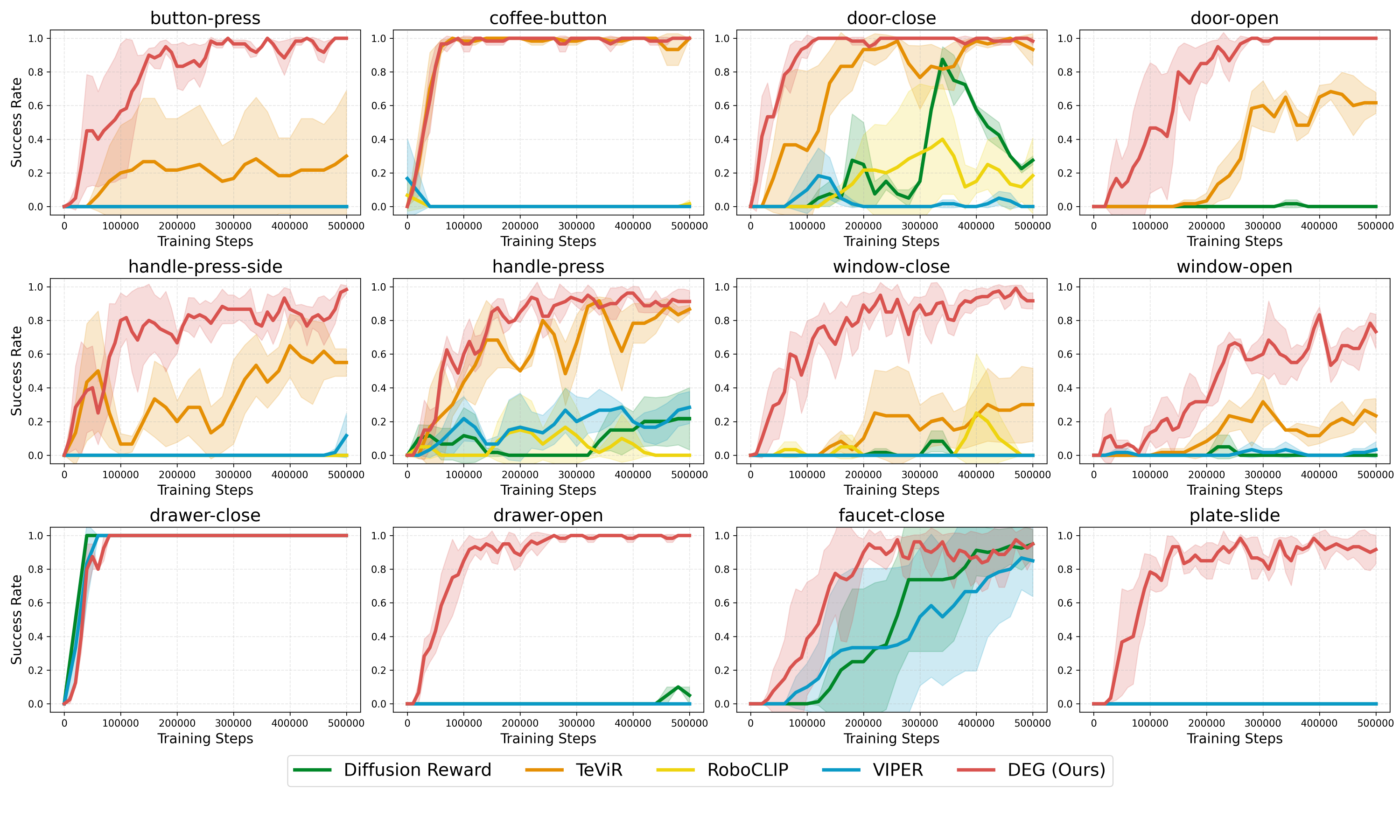}
    \vspace{-6mm}
    \caption{Comparison with state-of-the-art reward engineering methods on 12 task-free tasks. DEG achieves better performance on both the final policy level and RL sample efficiency. Owing to the non-open-source implementation or constraints on computational resources, we employ the results of TeViR and RoboCLIP provided by previous works \cite{tevir} only for the first eight tasks.}
    \label{reward-free-tasks}
\end{figure*}

where $I_{\text{reached}}$ is set to record the rearmost (i.e., largest index) frame that has been successfully matched so far, which prevents the agent from "slacking off" and repeatedly obtaining rewards by matching the same or previous frames. 

\paragraph{Dual-granularity reward} employed in DEG consists of the above two sub-incentives with coefficients $\alpha$ and $\beta$:

\begin{equation}
    \mathcal{R}_{DEG} = \alpha  \mathcal{R}_{coarse} + \beta\mathcal{R}_{fine}.
\end{equation}

It can guide effective RL independently or achieve more efficient learning with the help of a success sparse reward $\mathcal{R}_s$ that is easy to obtain and scaled by $\theta$:
\begin{equation}
    \mathcal{R}_{DEG+} = \alpha  \mathcal{R}_{coarse} + \beta\mathcal{R}_{fine} +\theta\mathcal{R}_s.
\end{equation}

\section{Experiment}

We conduct extensive experiments to answer the following questions: (Section 4.1) Can DEG, relying solely on a small amount of expert videos, independently guide effective RL and outperform SOTA baselines (equipped with sufficient multi-view expert data) in terms of sampling efficiency? (Section 4.2) With the aid of the sparse success reward, can DEG achieve performance comparable to that of handcrafted rewards designed by human experts? (Section 4.3) Can DEG also work in the real-world manipulation domains? (Section 4.4) Are both components of the dual-granularity reward effective? (Section 4.5) Compared with large-scale pre-trained encoders, does the contrastive self-supervised encoder possess superior latent space alignment capabilities? In addition, we further provide more results and analysis in Appendix A.

\begin{figure*}[t]
    \centering
    \includegraphics[width=0.91\textwidth]{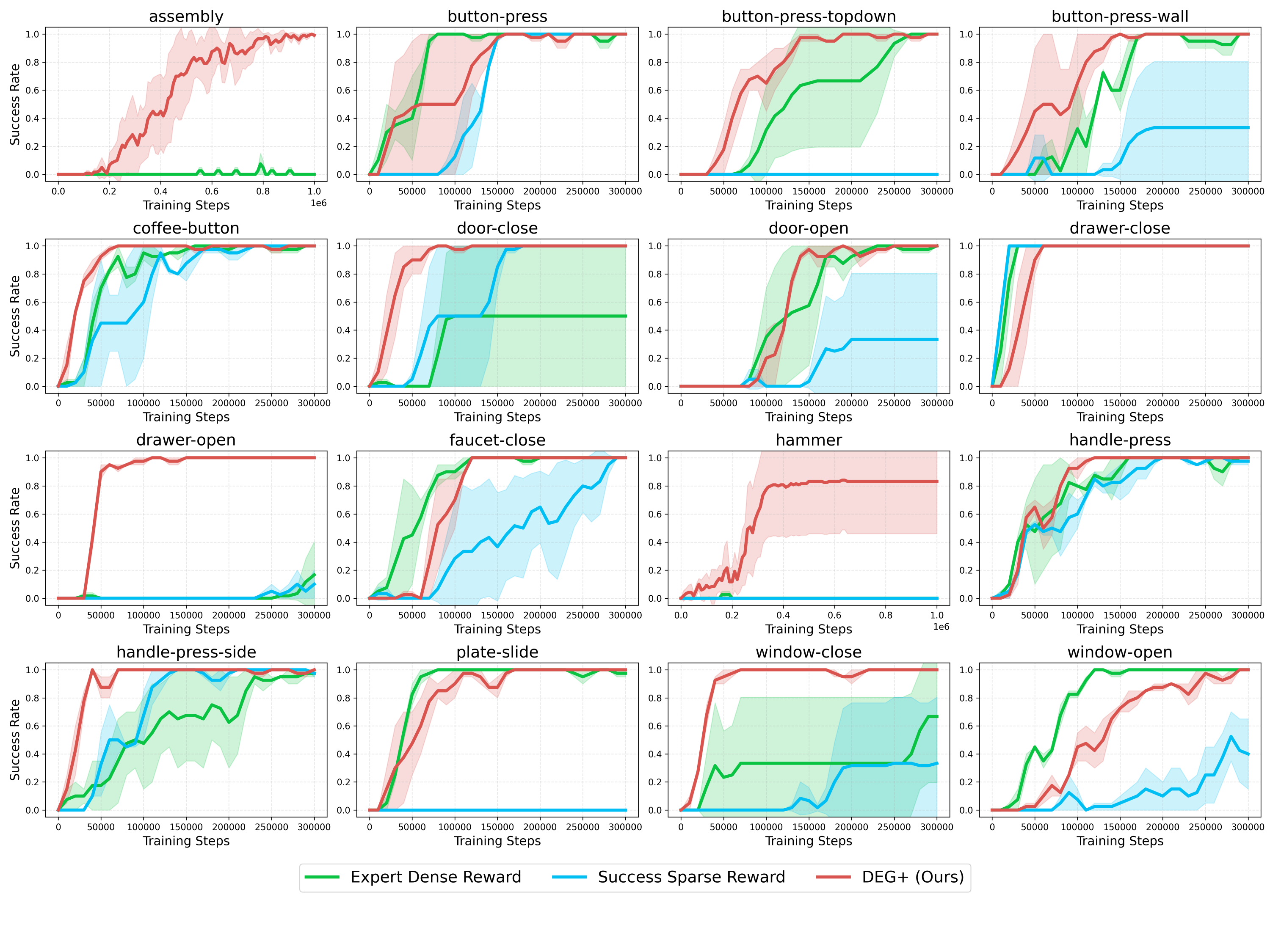}
    \vspace{-8mm}
    \caption{With success sparse reward, DEG+ can effectively improve RL efficiency and match human expert-annotated dense reward. It even outperforms expert dense reward across several tasks, such as \textit{assembly}, \textit{hammer}, and \textit{drawer-open}. }
    \label{16tasks}
\end{figure*}

\subsection{Comparison on Reward-Free Tasks}

In this section, we compare the proposed DEG on reward-free manipulation tasks against several popular and state-of-the-art baselines: TeViR \cite{tevir}, Diffusion Reward \cite{diffusionreward}, Viper \cite{viper}, and RoboCLIP \cite{roboclip}. Following these works, we conduct experiments on 12 challenging and diverse Metaworld manipulation tasks \cite{metaworld,metaworld-v3}. These tasks cover a diverse and challenging set of robotic arm operations, essentially spanning the entire feasible state space of the Sawyer robotic arm. DrQv2 \cite{drq-v2} is employed as the backbone for all methods, with only 500k steps allowed to fully evaluate the RL sample efficiency. See Appendix B for a detailed description of baseline methods and tasks.

For each task, DEG is only provided with 3-5 videos (details in Appendix B), and no other expert supervision or human annotations are used. Considering the different data requirements of the baselines, we conduct experiments according to their own settings or directly use the results from previous papers. For example, TeViR \cite{tevir} relies on 60 videos and requires three camera viewpoints (the left side of the arm, the top of the arm, and a close-up view of the gripper). We follow its requirements for TeViR, and the same applies to other methods. Note that although we present all results together for comparison, our DEG is still only trained with a few videos from a single camera.

The results are shown in Figure \ref{reward-free-tasks}. Except for \textit{door-close} where DEG and TeViR achieve similar final performance, and \textit{drawer-close} where all baselines converge well, DEG consistently and significantly outperforms all baselines on the remaining 10 tasks. This demonstrates DEG's ability to independently guide effective RL and ultimately drive stable policy convergence. Note that TeViR uses sufficient expert videos from three viewpoints, while DEG only uses a small number of single-view videos. In terms of sample efficiency, DEG also shows a huge advantage across all tasks, which proves the high precision of the proposed dual-granularity contrastive dense reward in the semantic guidance of robotic arm movements.

\subsection{DEG with Sparse Success Feedback}

In this section, we demonstrate the performance of DEG with the help of an easy-to-obtain success sparse reward, named DEG+. To demonstrate the effectiveness of DEG, we directly compare it with the latest human expert-annotated dense reward \cite{metaworld-v3} on the MetaWorld benchmark, and refer to this method as Expert Dense Reward. In addition, the RL results of Success Sparse Reward \cite{metaworld} are also used for comparison. 

The results across all 16 tasks are shown in Figure \ref{16tasks}. Without dense incentives, Success Sparse Reward struggles to guide stable RL. This is because the exploration space in long-horizon tasks is vast, and it is hard for the agent to complete the task and obtain signals through the exploration provided by the randomness of the stochastic policy. In contrast, DEG provides an exclusive successful trajectory for each RL episode and delivers efficient guiding rewards accordingly, which greatly improves the efficiency of exploring sparse success feedback. In addition, even when compared with the dense rewards handcrafted by human experts, DEG can still match their performance on most tasks and outperform them on several tasks, mainly due to the complete information provided by expert guidance videos. Take \textit{drawer-open} as an example: the expert dense reward incentivizes the distance between the gripper and the handle, which can cause the gripper to push against the handle from the outside, making it hard to open the drawer, falling into a local optimum. DEG, however, can accurately learn from the generated expert guidance to lift the gripper, aim at the inner side of the handle, and then lower it, achieving extremely stable policy improvement.

\subsection{Real-World Experiments}

\begin{figure*}[t]
    \centering
    \includegraphics[width=0.95\textwidth]{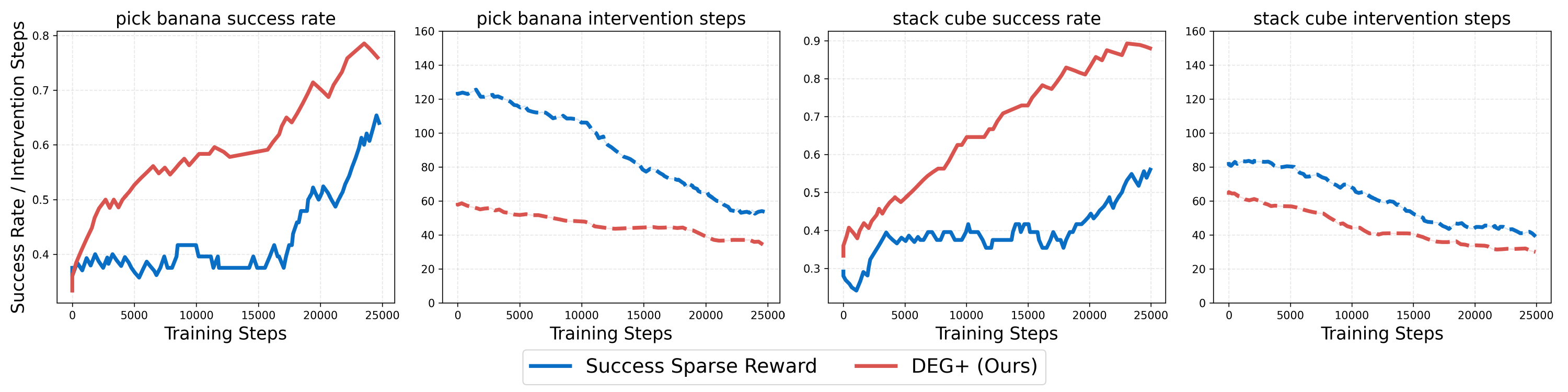}
    
    \caption{Smoothed training curves on real-world Franka manipulation tasks. DEG enables better RL efficiency and lower intervention rates than employing only the success sparse reward.}
    \vspace{-5mm}
    \label{fig:franka}
    \vspace{-1mm}
\end{figure*}

To further explore the capabilities of DEG, we conduct experiments on physical robotic arm manipulation tasks. We design two tasks using the Franka robotic arm: \textit{stack-cube} and \textit{pick-banana}. Both are long-horizon tasks that require precise interaction between the robotic arm and objects. You can refer to Appendix B for a more detailed description of the real-world tasks. We adopt the state-of-the-art human-in-the-loop RL algorithm, HIL-SERL \cite{hilserl}, as the backbone of the physical experiments and allow a maximum of 25k steps and 3 expert videos for each task. 

The results are shown in Figure \ref{fig:franka}. DEG can also provide real-world robotic manipulation with effective task-level dense incentives, significantly improving the sampling efficiency of real-world RL. In addition, we observe that DEG reward is triggered with extremely high frequency when humans provide help. This indicates that the DEG reward can effectively align with expert knowledge, thereby improving the efficiency of learning useful demonstrations and reducing the intervention rate. Since object states are not directly accessible in the real world as they are in simulators, handcrafting dense rewards becomes extremely difficult, while DEG can realize effective dense reward design for physical tasks with very easy data collection.

\begin{figure}[t]
    \centering
    \includegraphics[width=0.48\textwidth]{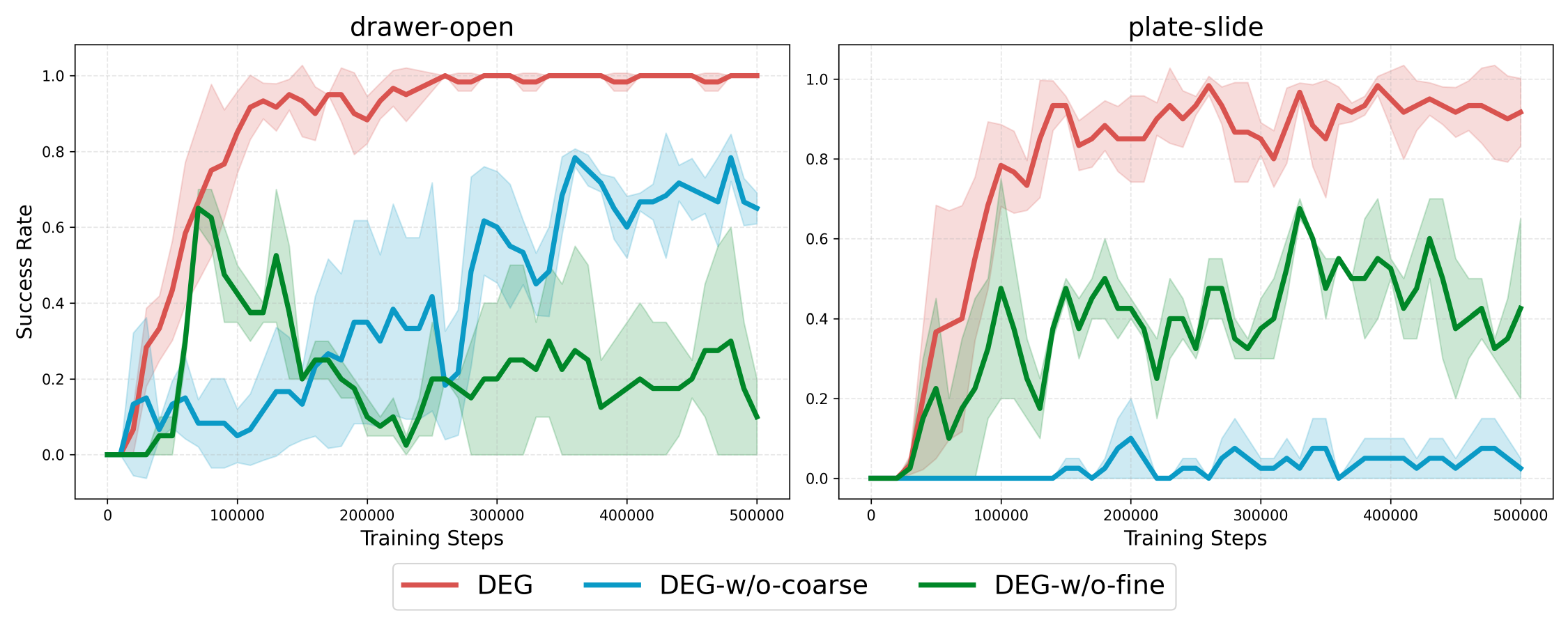}
    
    \caption{Numerical ablation experiments. Both the coarse-grained reward and fine-grained reward are useful in DEG. }
    \label{ablation-main}
    \vspace{-5mm}
\end{figure}

\subsection{Ablation Study}

In this section, we provide ablation experiments on the proposed dual-granularity reward, verifying whether both sub-rewards are useful. Experimental results (Figure \ref{ablation-main}) confirm that both contribute to DEG's achievement of the sample-efficient RL. In addition, we further provide a very intuitive visualization of the policy intentions learned by DEG and its two ablation variants on the \textit{drawer-open} task, shown in Figure \ref{ablation-visualization} in Appendix A. With only the coarse-grained reward, DEG can quickly learn the overall motion trajectory of the robotic arm in the expert videos but fails to achieve precise interaction with the drawer. With only the fine-grained reward, the robotic arm can grasp the handle yet lacks the tendency to imitate the motion trajectory. In contrast, the dual-granularity reward effectively combines the strengths of both, enabling correct handle interaction and smooth task completion.

\begin{figure}[t]
  \centering
  \subfloat{
    \includegraphics[width=0.22\textwidth]{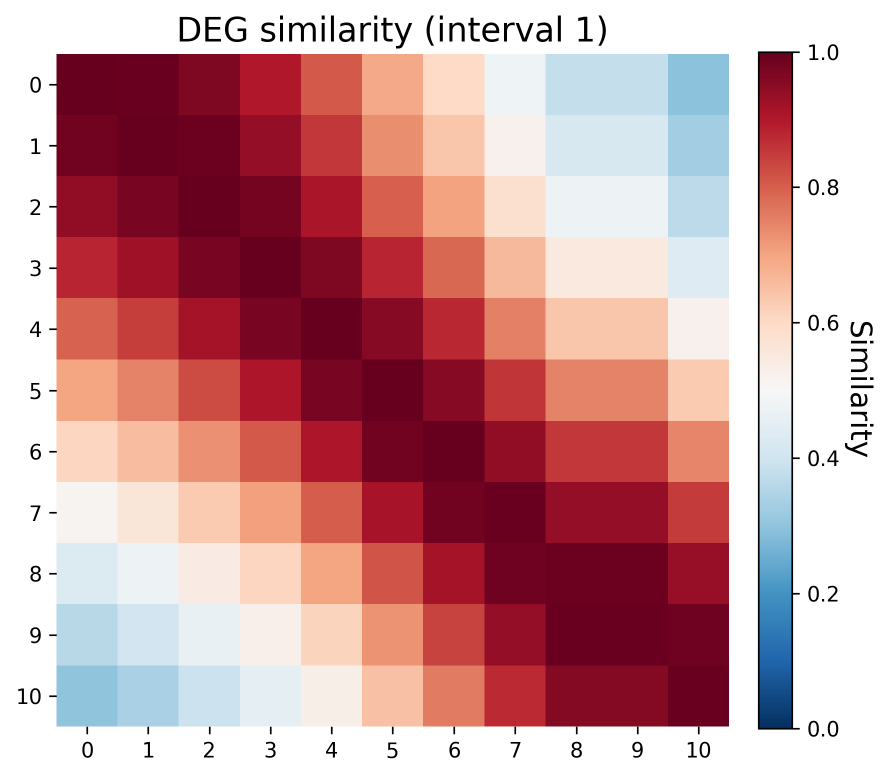}
  }
  \quad
  \subfloat{
    \includegraphics[width=0.22\textwidth]{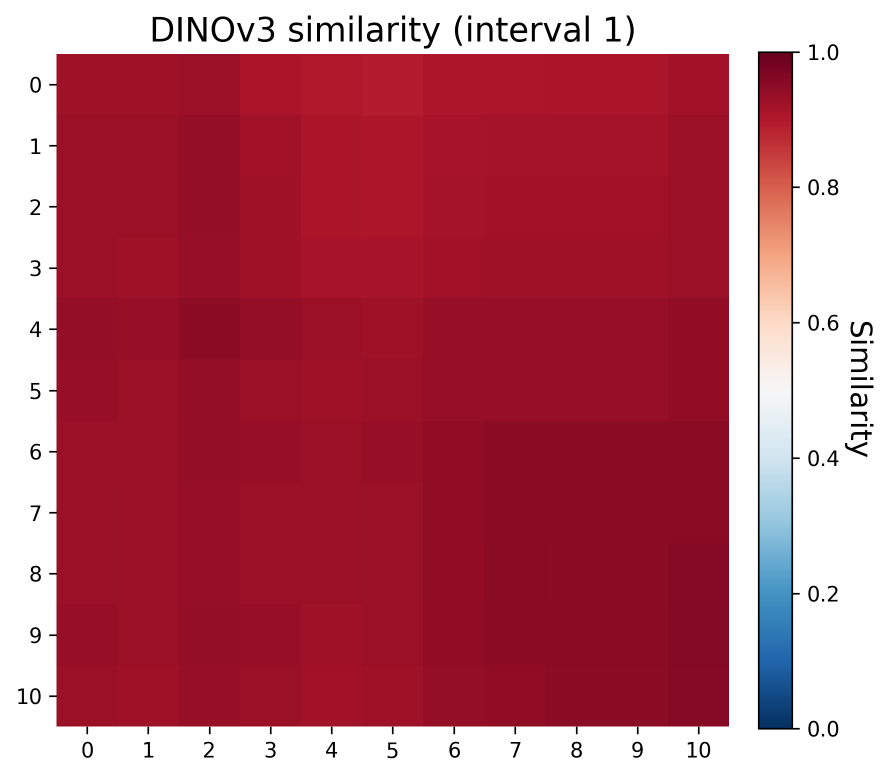}
  }
  \caption{Heatmaps from two different encoders: frame similarity between generated videos and real observations with identical trajectories (\textit{drawer-open}). The vertical axis represents the real observation sequence (sampled at an interval of 1), while the horizontal axis represents the corresponding generated frames. Compared with DINOv3 (right), our DEG contrastive encoder (left) can (i) better align frames with the same semantics while distinguishing those with different semantics and (ii) map the semantic distance into a proper similarity distance. 
  }
  \vspace{-6mm}
  \label{heatmap}
\end{figure}

\subsection{Encoder Analysis}

In DEG, we train the reward calculation encoder via contrastive self-supervised learning, instead of adopting state-of-the-art pre-trained models such as DINOv3 \cite{dinov3}. In this section, we visualize the matching performance of different encoders when facing real and generated videos, as shown in Figure \ref{heatmap}. The pre-trained DINOv3 can adapt to different domains via downstream predictors, while its extensive prior knowledge makes minor visual variations in a single target domain less distinguishable in the latent space. By contrast, the DEG contrastive encoder can well align semantically identical frames and distinguish semantically different ones. In addition, the contrastive encoder can also map the semantic distance into a proper similarity change in the latent space, thus enabling DEG to design dual-granularity rewards based on this similarity. We provide more visualizations in Appendix A.

\section{Conclusion \& Limitation}

In this paper, we propose DEG, a dense reward framework for sample-efficient embodied RL without requirements on human annotations or extensive supervisions. By employing a finetuned video generation model as an RL guide, the proposed dual-granularity reward effectively encourages the agent to imitate the generated guidance in the self-supervised space, achieving efficient RL that surpasses state-of-the-art baselines. The main bottleneck of DEG lies in the slow video generation speed. Since pre-training data for video generation models is relatively high-fidelity, reducing the resolution to speed up generation may cause collapse. We believe this issue will likely be resolved with the development of foundation models, and DEG can inspire more researchers to focus on data-efficient dense reward design.

\section*{Impact Statement}
This paper presents work whose goal is to advance the field of 
Machine Learning. There are many potential societal consequences 
of our work, none which we feel must be specifically highlighted here.


\nocite{langley00}

\bibliography{example_paper}
\bibliographystyle{icml2026}

\newpage
\appendix
\onecolumn

\section{Additional Results \& Analysis}
\subsection{More Ablation Results}

\begin{figure}[h]
    \centering
    \includegraphics[width=0.95\textwidth]{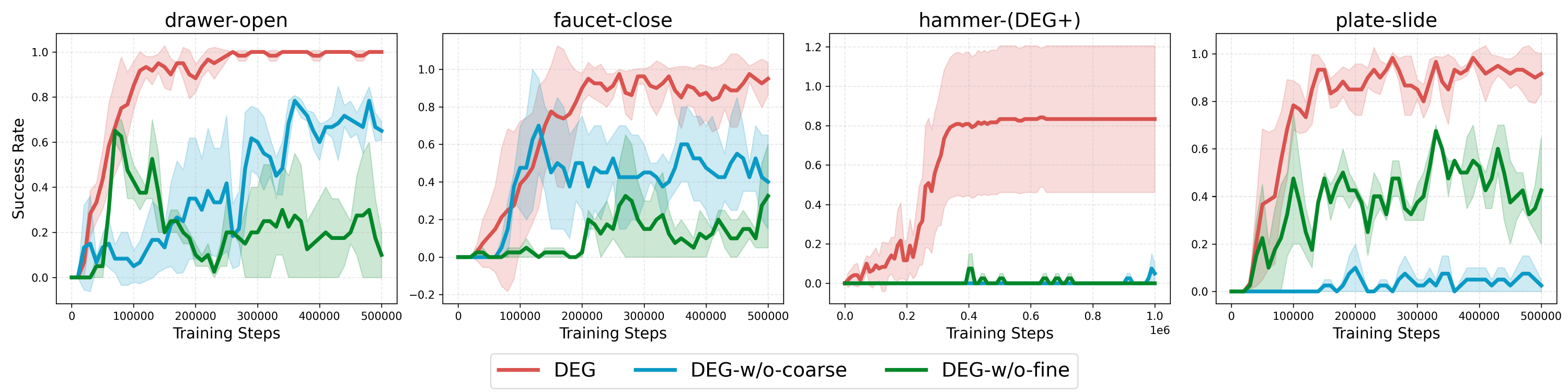}
    
    \caption{Numerical ablation study on three reward-free tasks and one sparse-reward task (marked with DEG+). Both coarse-grained and fine-grained rewards are necessary.}
    \label{ablation1-4}
\end{figure}

In addition to the results in Section 4.4, we provide more ablation experiments on three reward-free tasks and one sparse-reward task (marked with DEG+). The results in Figure \ref{ablation1-4} further demonstrate that both sub-rewards are necessary in the proposed dual-granularity reward.

\subsection{Ablation: Policy Intention Visualization}

\begin{figure}[h]
    \centering
    \includegraphics[width=0.95\textwidth]{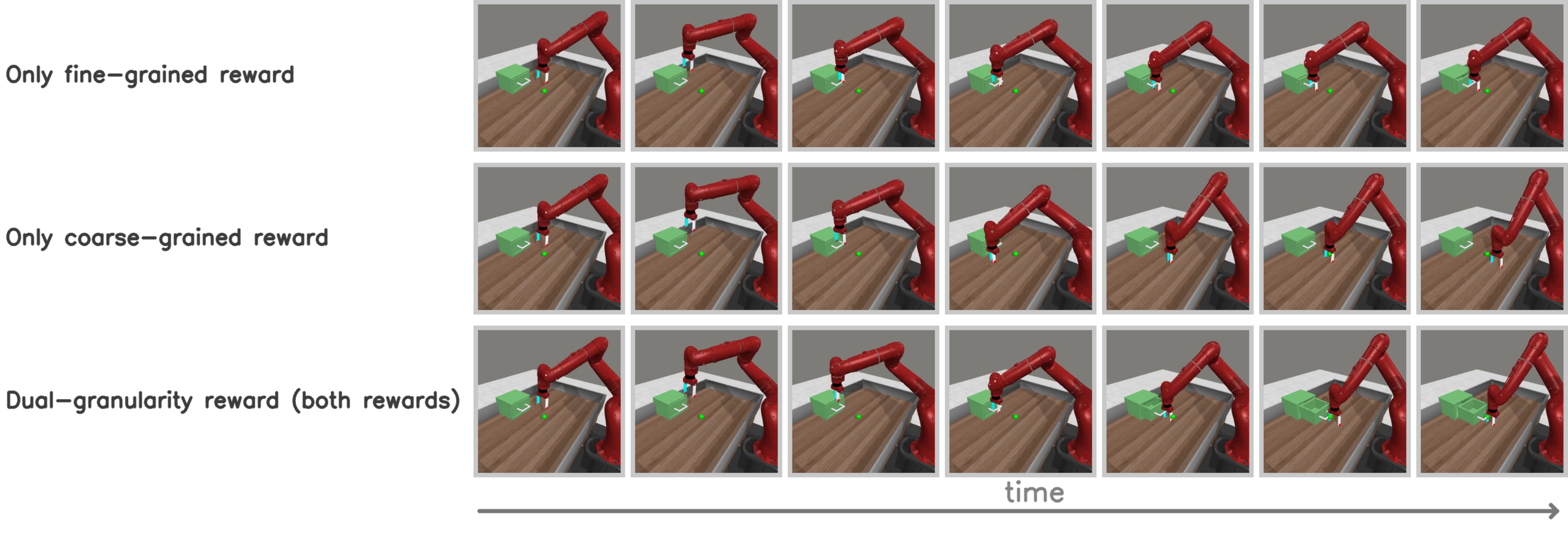}
    
    \caption{Policy intention visualization of DEG and its two ablation variants. Each of the two sub-rewards has its own strengths and weaknesses, while the dual-granularity reward can combine the advantages of both.}
    \label{ablation-visualization}
\end{figure}

In addition to the numerical comparison, we further provide the visualization of the policy intentions learned by DEG and its two ablation variants. We employ the learned policy in the reward-free \textit{drawer-open} task at 250k environmental steps and show the results in Figure \ref{ablation-visualization}. With only the coarse-grained reward, DEG can quickly learn the overall motion trajectory of the robotic arm (contained in the expert videos) but fails to achieve precise interaction with the drawer. With only the fine-grained reward, the robotic arm can grasp the handle yet lacks the tendency to imitate the motion trajectory. In contrast, the dual-granularity reward effectively combines the strengths of both, enabling precise and smooth task completion.

\newpage
\section{More Encoder Visualization and Analysis}

\begin{figure}[h]
  \centering
  \subfloat{
    \includegraphics[width=0.25\textwidth]{images/dinov3-1-output.pdf}
  }
  \quad
  \subfloat{
    \includegraphics[width=0.25\textwidth]{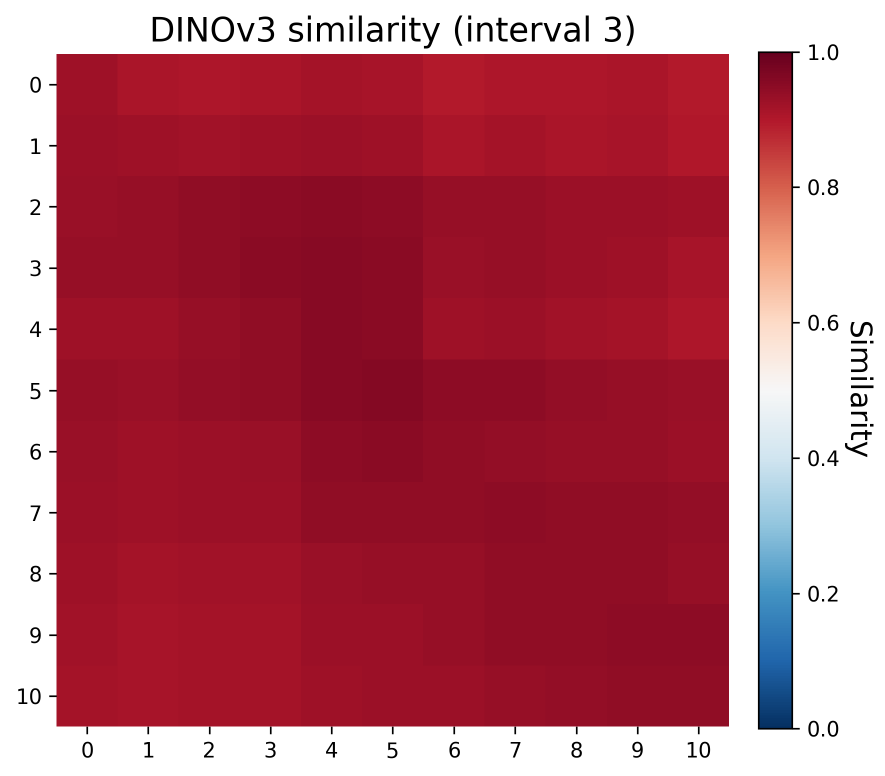}
  }
  \quad
  \subfloat{
    \includegraphics[width=0.25\textwidth]{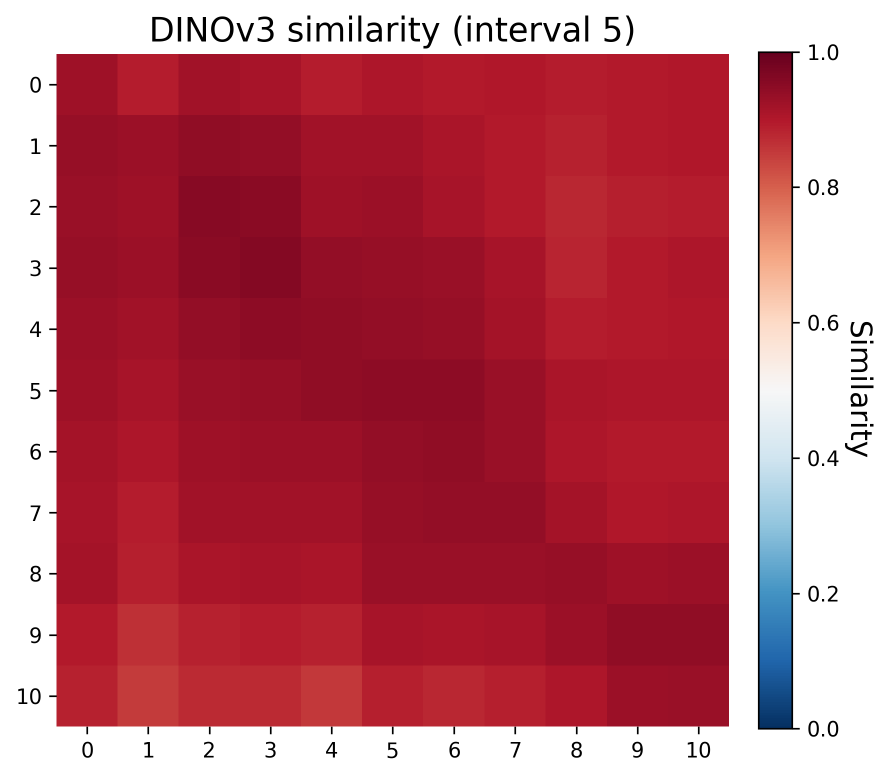}
  }
  \caption{Frame similarity between generated videos and real observations (with intervals 1, 3, and 5) with identical semantics (\textit{drawer-open}) when using DINOv3. The vertical axis represents the real observation sequence (sampled at an interval of 1, 3, and 5), while the horizontal axis represents the corresponding generated frames. Specifically, an interval of 1 means the frame sequence used for visualization is frames indexed 0, 1, 2, ..., 10 from the real video; an interval of 3 denotes using frames indexed 0, 3, 6, ..., 27 from the real video; and an interval of 5 denotes using frames indexed 0, 5, 10, ..., 50 from the real video. The color distributions remain highly similar across different intervals, indicating that DINOv3 cannot adequately map semantic distances to similarity differences. 
  }
  
  \label{dinov3-map-all}
\end{figure}

In addition to the comparison in Section 4.5, we further provide more results and analysis on the DEG contrastive encoder and pre-trained DINOv3 \cite{dinov3}. We first present the performance of DINOv3 on video sequences with different frame intervals, as shown in Figure \ref{dinov3-map-all}. DINOv3 exhibits slightly increased discrimination for frames with larger intervals (i.e., larger semantic distances), but the color distributions remain highly similar, indicating that it cannot adequately map semantic distances to similarity differences, which makes it unsuitable for similarity-based reward design.

\begin{figure}[h]
  \centering
  \subfloat{
    \includegraphics[width=0.25\textwidth]{images/ours-1-output.pdf}
  }
  \quad
  \subfloat{
    \includegraphics[width=0.25\textwidth]{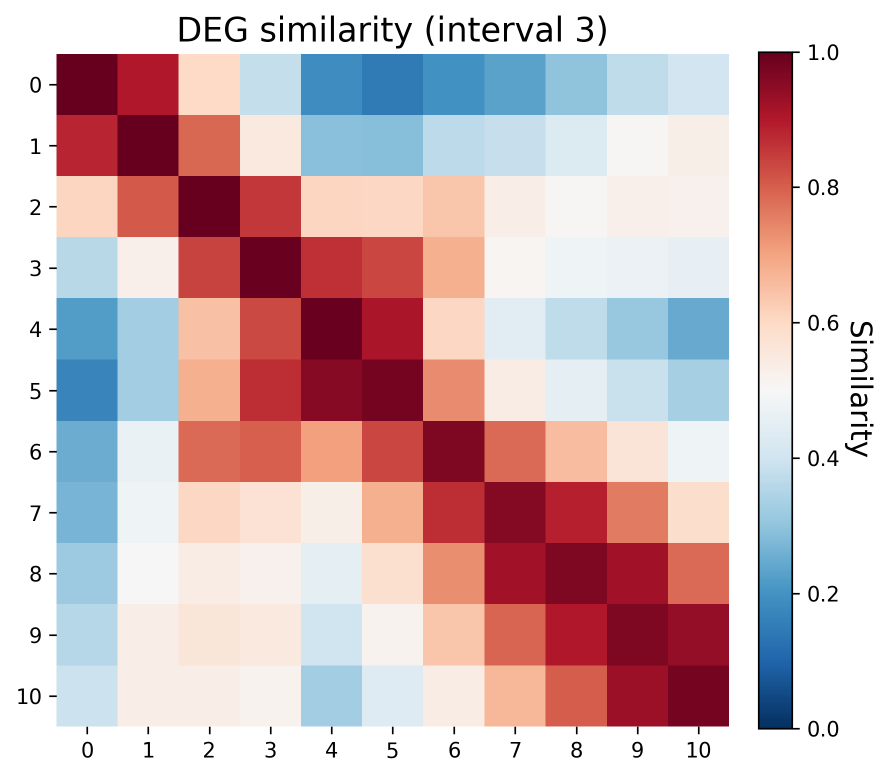}
  }
  \quad
  \subfloat{
    \includegraphics[width=0.25\textwidth]{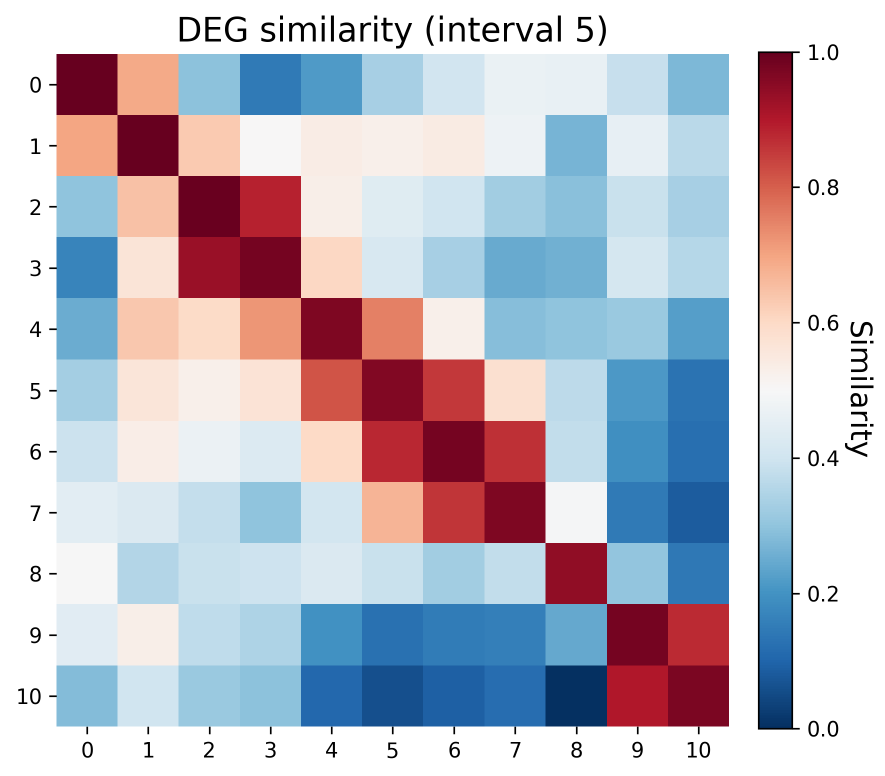}
  }
  \caption{Frame similarity between generated videos and real observations (with intervals 1, 3, and 5) with identical semantics (\textit{drawer-open}) when using our DEG contrastive encoder. The vertical axis represents the real observation sequence (sampled at an interval of 1, 3, and 5), while the horizontal axis represents the corresponding generated frames. Specifically, an interval of 1 means the frame sequence used for visualization is frames indexed 0, 1, 2, ..., 10 from the real video; an interval of 3 denotes using frames indexed 0, 3, 6, ..., 27 from the real video; and an interval of 5 denotes using frames indexed 0, 5, 10, ..., 50 from the real video. The DEG contrastive encoder shows significantly stronger discrimination for frame sequences with larger intervals, while maintaining the ability to align semantically identical frames over a wider temporal range. 
  }
  
  \label{DEG-map-all}
\end{figure}

In contrast, the DEG contrastive encoder (Figure \ref{DEG-map-all}) shows significantly stronger discrimination for frame sequences with larger intervals, while maintaining the ability to align semantically identical frames over a wider temporal range. Furthermore, for frames that are semantically close but distinct, DEG can effectively map their semantic distance to the similarity difference (heatmap of interval 1), thereby supporting the design of the proposed dual-granularity reward with different thresholds.

\newpage

\subsection{Generated Episodic Guidance in DEG}

\begin{figure}[h]
    \centering
    \includegraphics[width=0.95\columnwidth]{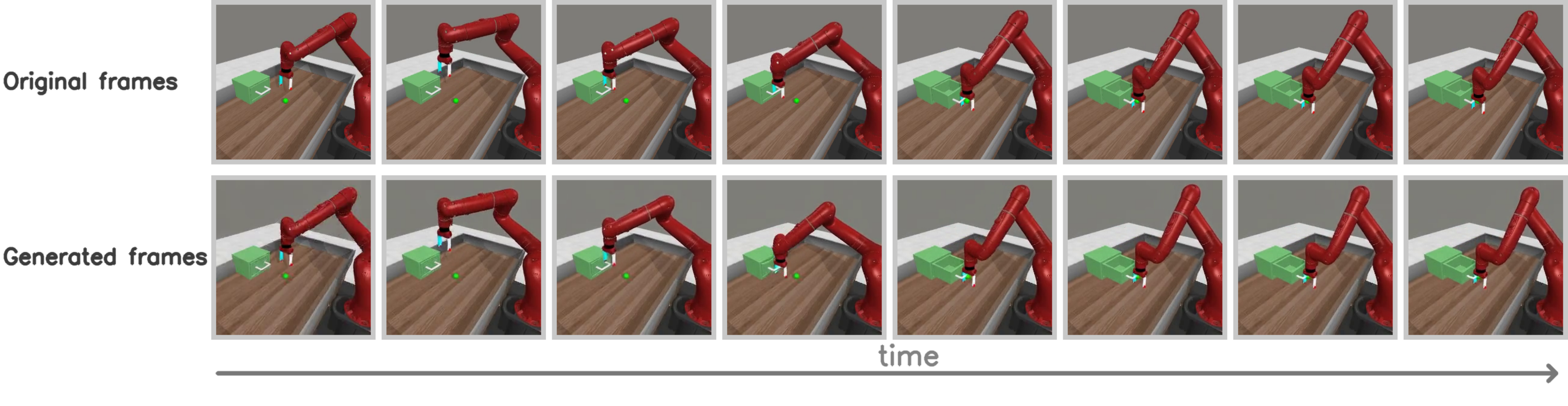}
    \caption{DEG generated videos versus real videos on Metaworld \textit{drawer-open} task.}
    \label{fig:pull_drawer}
\end{figure}

\begin{figure}[h]
    \centering
    \includegraphics[width=0.95\columnwidth]{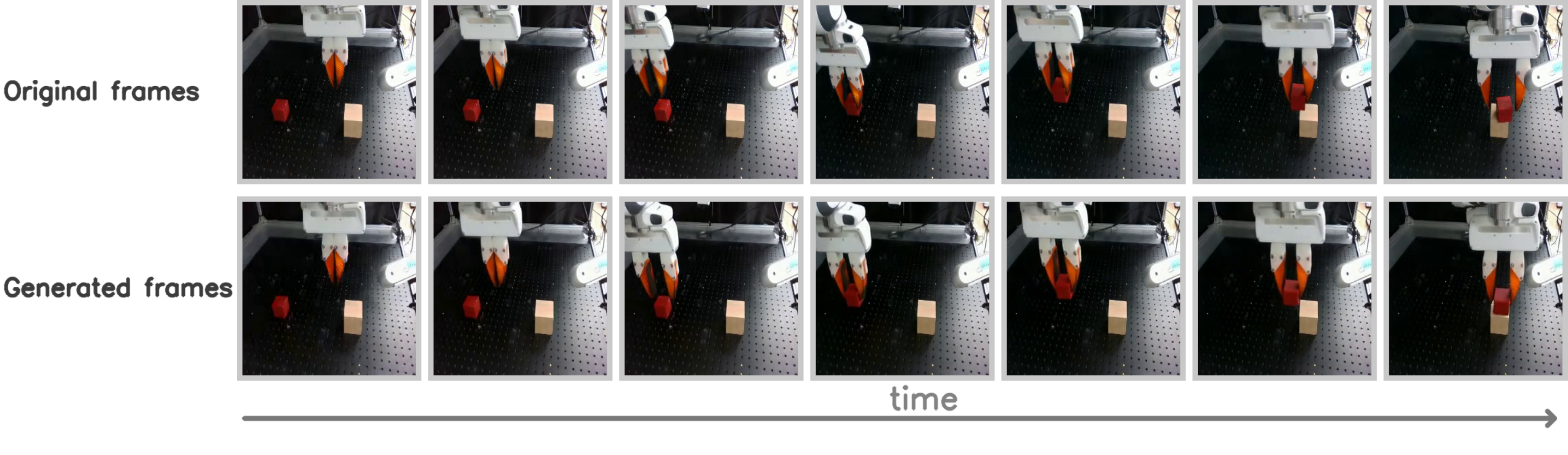}
    \caption{DEG generated videos versus real videos on real-world \textit{stack-cube} task.}
    \label{fig:stack_cube}
\end{figure}

In this chapter, we present a direct comparison between the real videos and generated videos, shown in Figure \ref{fig:pull_drawer} and Figure \ref{fig:stack_cube}. We find that the fine-tuned video generation model can achieve effective domain adaptation. Despite remaining noise in terms of clarity and certain details, it basically recovers the semantic manner of real task execution. Note that real videos (3-5 clips) are only used to fine-tune the video generation model, and DEG uses generated videos to calculate rewards throughout the entire RL process.

\newpage
\section{Experimental Details}

\subsection{Task Description} 

Following previous works \cite{diffusionreward,tevir}, we employ sixteen simulation tasks from Metaworld \cite{metaworld,metaworld-v3}: \textit{door-close},\textit{window-close}, \textit{handle-press}, \textit{button-press}, \textit{coffee-button}, \textit{drawer-open}, \textit{button-press-wall}, \textit{drawer-close}, \textit{plate-slide}, \textit{button-press-topdown}, \textit{hammer}, \textit{assembly},\textit{door-open},  \textit{window-open}, \textit{handle-press-side}, \textit{faucet-close}.
These tasks cover a diverse and challenging set of robotic arm operations, essentially spanning the entire feasible state space of the Sawyer robotic arm.  For real-world tasks, we employ the Franka arm as the physical manipulator, designing two tasks: \textit{pick-banana} and \textit{stack-cube}. We provide the description of the above 18 tasks below:

\begin{figure}[H]
\centering
\begin{tabular}{@{}m{0.13\columnwidth} m{0.3\columnwidth} m{0.13\columnwidth} m{0.3\columnwidth}@{}}
\includegraphics[width=\linewidth]{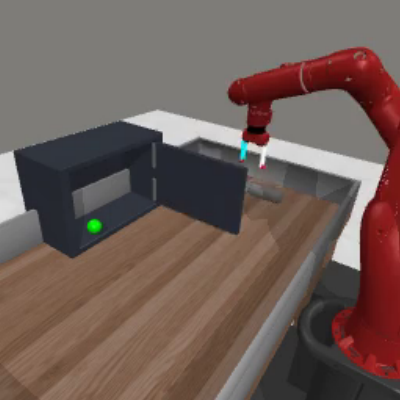}
&
{\raggedright\small
\textit{door-close.}\\
Closing a door mounted on a revolving hinge joint. The task requires adapting the pushing trajectory to account for the door’s randomized locations.
}
&
\includegraphics[width=\linewidth]{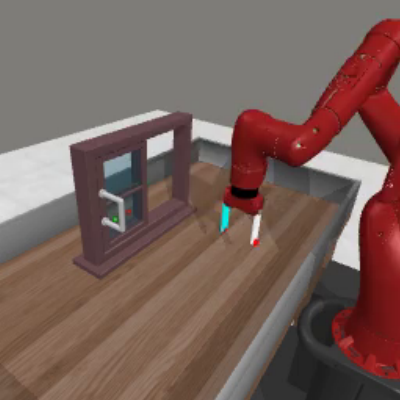}
&
{\raggedright\small
\textit{window-close.}\\
Executing a push motion to close a sliding window with a randomized position.
}
\end{tabular}
\end{figure}
\begin{figure}[H]
\centering
\begin{tabular}{@{}m{0.13\columnwidth} m{0.3\columnwidth} m{0.13\columnwidth} m{0.3\columnwidth}@{}}
\includegraphics[width=\linewidth]{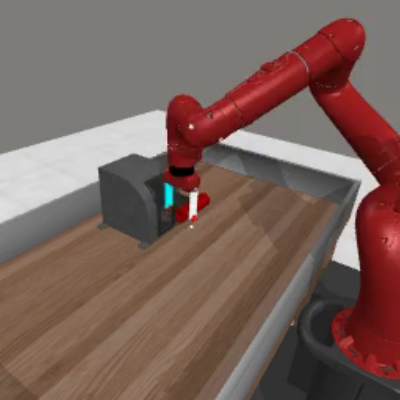}
&
{\raggedright\small
\textit{handle-press.}\\
Depressing a lever handle to its endpoint. The handle's initial locations are varied.
}
&
\includegraphics[width=\linewidth]{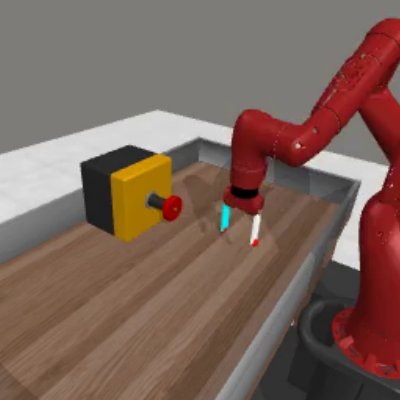}
&
{\raggedright\small
\textit{button-press.}\\
 Actuating a standard push-button from a randomized location in the workspace.
}
\end{tabular}
\end{figure}
\begin{figure}[H]
\centering
\begin{tabular}{@{}m{0.13\columnwidth} m{0.3\columnwidth} m{0.13\columnwidth} m{0.3\columnwidth}@{}}
\includegraphics[width=\linewidth]{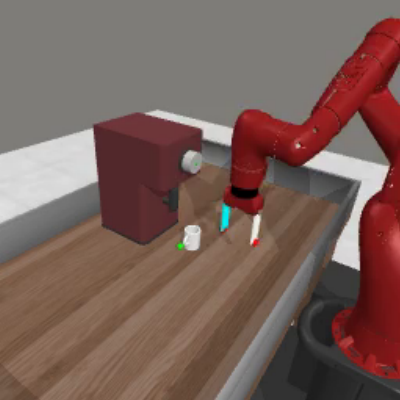}
&
{\raggedright\small
\textit{coffee-button.}\\
Pushing a button on a coffee machine to trigger dispensing.
The position of the coffee machine is randomized.
}
&
\includegraphics[width=\linewidth]{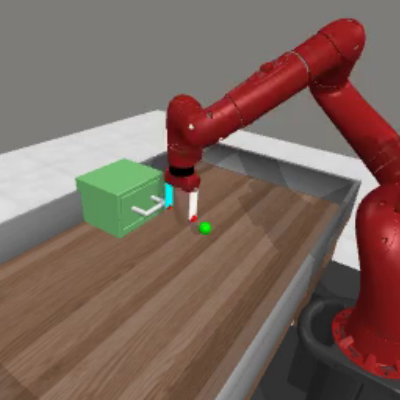}
&
{\raggedright\small
\textit{drawer-open.}\\
 Grasping and pulling open a drawer from a randomized closed position.
}
\end{tabular}
\end{figure}
\begin{figure}[H]
\centering
\begin{tabular}{@{}m{0.13\columnwidth} m{0.3\columnwidth} m{0.13\columnwidth} m{0.3\columnwidth}@{}}
\includegraphics[width=\linewidth]{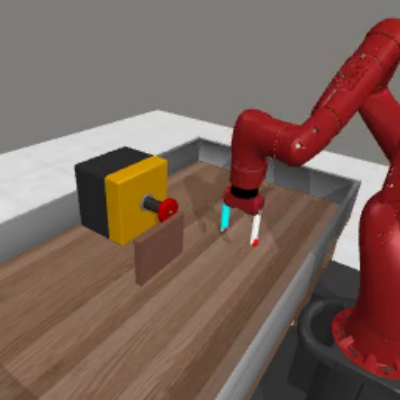}
&
{\raggedright\small
\textit{button-press-wall.}\\
A navigation and manipulation task that involves bypassing a physical barrier to reach and press a button located behind it, with randomized relative positioning.
}
&
\includegraphics[width=\linewidth]{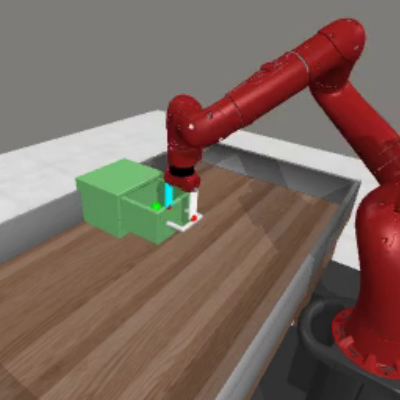}
&
{\raggedright\small
\textit{drawer-close.}\\
 Pushing a drawer shut from a randomized open displacement.
}
\end{tabular}
\end{figure}
\begin{figure}[H]
\centering
\begin{tabular}{@{}m{0.13\columnwidth} m{0.3\columnwidth} m{0.13\columnwidth} m{0.3\columnwidth}@{}}
\includegraphics[width=\linewidth]{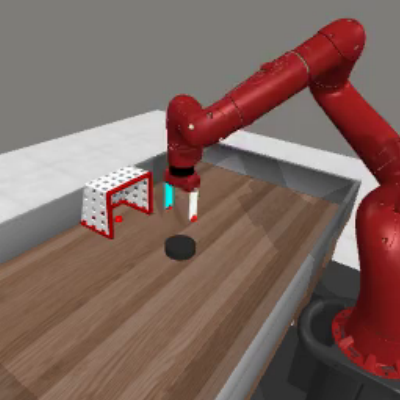}
&
{\raggedright\small
\textit{plate-slide.}\\
Sliding a plate laterally into a constrained cabinet opening. The initial placements of the target cabinet are varied.
}
&
\includegraphics[width=\linewidth]{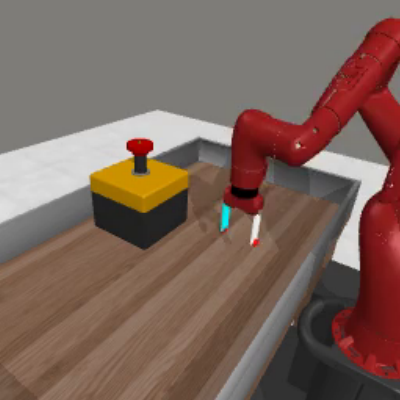}
&
{\raggedright\small
\textit{button-press-topdown.}\\
 Pressing a button via a precise top-down approach, with randomized button placement.
}
\end{tabular}
\end{figure}
\begin{figure}[H]
\centering
\begin{tabular}{@{}m{0.13\columnwidth} m{0.3\columnwidth} m{0.13\columnwidth} m{0.3\columnwidth}@{}}
\includegraphics[width=\linewidth]{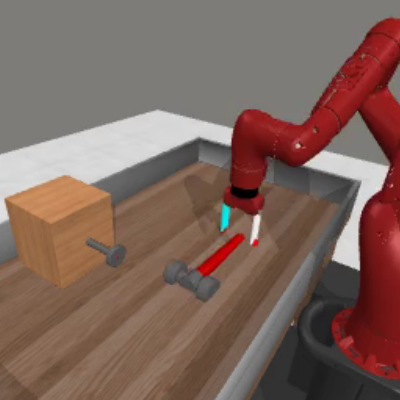}
&
{\raggedright\small
\textit{hammer.}\\
A dynamic task requiring the accurate swinging of a hammer to strike the head of a wall-mounted screw. Randomization of the tool positions assesses swing trajectory planning and impact control.
}
&
\includegraphics[width=\linewidth]{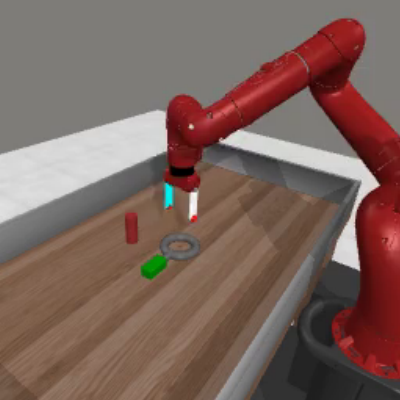}
&
{\raggedright\small
\textit{assembly.}\\
 A bi-manipulation task involving the sequential picking of a free nut and its precise placement onto a fixed peg. Randomization of the peg positions tests fine grasping, alignment, and insertion skills.
}
\end{tabular}
\end{figure}
\begin{figure}[H]
\centering
\begin{tabular}{@{}m{0.13\columnwidth} m{0.3\columnwidth} m{0.13\columnwidth} m{0.3\columnwidth}@{}}
\includegraphics[width=\linewidth]{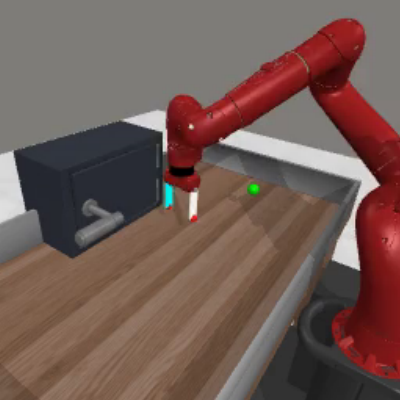}
&
{\raggedright\small
\textit{door-open.}\\
Opening a hinged door from a randomized closed position, involving grasping and pulling a handle from variable configurations.
}
&
\includegraphics[width=\linewidth]{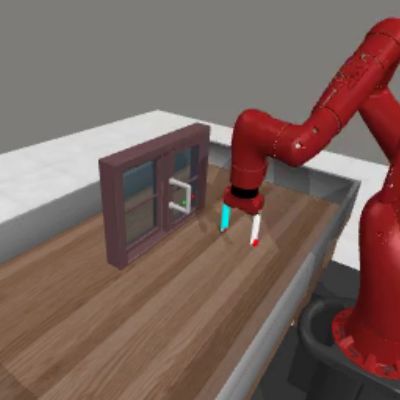}
&
{\raggedright\small
\textit{window-open.}\\
 Opening a sliding window by pushing it from a randomized initial closed state.
}
\end{tabular}
\end{figure}
\begin{figure}[H]
\centering
\begin{tabular}{@{}m{0.13\columnwidth} m{0.3\columnwidth} m{0.13\columnwidth} m{0.3\columnwidth}@{}}
\includegraphics[width=\linewidth]{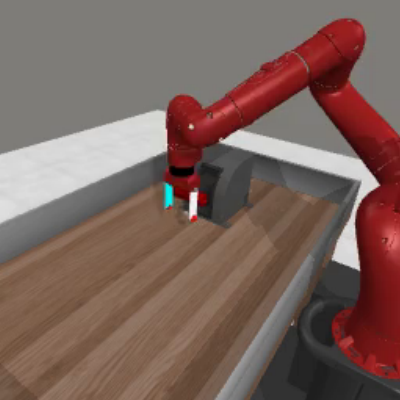}
&
{\raggedright\small
\textit{handle-press-side.}\\
Applying downward force on a handle along a sideways axis, with randomized handle poses.
}
&
\includegraphics[width=\linewidth]{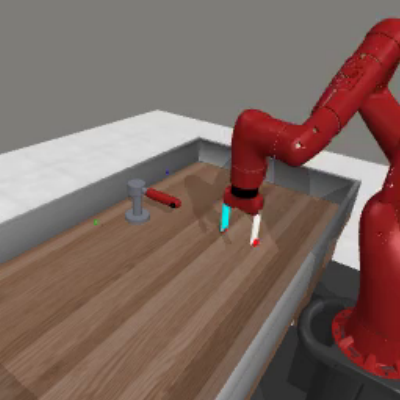}
&
{\raggedright\small
\textit{faucet-close.}\\
 Rotating a faucet knob clockwise to its closed position, starting from a randomized angular position.
}
\end{tabular}
\end{figure}

\begin{figure}[H]
\centering
\begin{tabular}{@{}m{0.13\columnwidth} m{0.3\columnwidth} m{0.13\columnwidth} m{0.3\columnwidth}@{}}
\includegraphics[width=\linewidth]{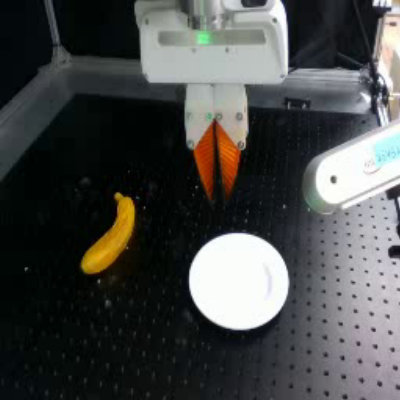}
&
{\raggedright\small
\textit{Real-world: pick-banana.}\\
Grasping a banana from a randomized initial position and placing it onto a fixed-location white plate. The task evaluates precision grasping of deformable objects and targeted placement.
}
&
\includegraphics[width=\linewidth]{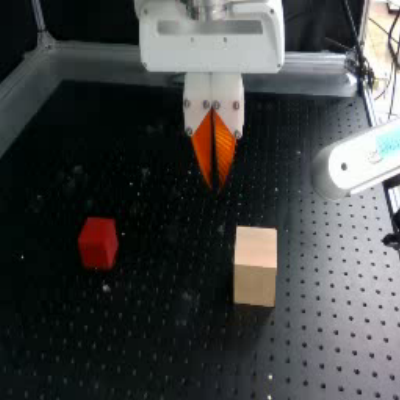}
&
{\raggedright\small
\textit{Real-world: stack-cube.}\\
 Stacking a small red cube from a randomized start position onto a larger, fixed-position natural wood-colored cube. The task assesses spatial alignment and stable placement for basic assembly.
}
\end{tabular}
\end{figure}

\subsection{Baseline Introduction} 

In Section 4.1, we employ four popular and advanced baselines: TeViR \cite{tevir}, Diffusion Reward \cite{diffusionreward}, VIPER \cite{viper}, and RoboCLIP \cite{roboclip}. 

\textbf{TeViR} \cite{tevir} utilizes a conditional video diffusion model to generate the multi-view trajectory of distance-based reward calculation. It is trained under three different cameras.

\textbf{Diffusion Reward} \cite{diffusionreward} leverages a conditional video diffusion model to capture the expert demonstration distribution and uses the conditional entropy of the video prediction with the sparse reward to formulate dense rewards.

\textbf{VIPER} \cite{viper} utilizes a video prediction transformer and directly leverages the model’s prediction likelihood as a reward signal to encourage the agent to learn the expected behavior. 

\textbf{RoboCLIP} \cite{roboclip} utilizes pre-trained Vision-Language Models (VLM) to generate rewards by calculating the similarity of the agent’s observation to the language task description for the agent.

In Section 4.2, we use the Success Sparse Reward and the Expert Dense Reward provided by the Metaworld environment \cite{metaworld,metaworld-v3}.

\textbf{Success Sparse Reward} \cite{metaworld} corresponds to the success signals provided by the environments. It is proportional to info['success'] in standard Metaworld environments.

\textbf{Expert Dense Reward} \cite{metaworld-v3} is the latest dense reward designed by experts on the Metaworld benchmark. This reward ensures solvability with PPO \cite{ppo} while being minimally opinionated.


\newpage
\subsection{Hyper-Parameters}

We provide the hyper-parameters of DEG, as shown in \ref{hyper}. More information (prompts and detailed number of expert videos) about video generation model finetuning is provided in the next section.

\begin{table}[h]
\caption{ The hyper-parameters of DEG. For domain \textit{door}, \textit{coffee}, task \textit{drawer-close}, and task \textit{button-press}, ($I_{coarse}$, $I_{fine}$) is (0.8, 0.95); for domain \textit{window} and \textit{hammer}, ($I_{coarse}$, $I_{fine}$) is (0.8, 0.96); for \textit{faucet}, $I_{coarse}$ is 0.9; and for \textit{assembly}, $s$ is 2 while $\theta$ is 100. For the detailed number of used videos, please refer to the next section. }
\centering
\setlength{\tabcolsep}{3mm}{
\begin{tabular}{|lc|}
\hline
Hyper-parameter                      & Setting       \\ \hline
Video fondation Model & Wan2.1-I2V-14B \\
Number of finetuning videos &  3-5 (details in next section)\\
Video frames & $250$ \\
Frame shape & $256\times256\times3$ \\
Frame stack & $1$\\
Action repeat & $2$\\
Action type & Continuous \\
Action dimension & See \cite{metaworld-v3}\\
Large model finetuning & LoRA\\

self-supervised pre-training times & $5000$ \\
Feature encoder lr & $1e-4$ \\
Random shift (noise) upper bound& $4$ \\
EMA update frequency & $2$ \\
EMA Momentum & 0.05 \\
Optimizer & Adam \cite{adam} \\

Coarse-grained coefficient $\alpha$& 100 \\
Coarse-grained threshold $I_{coarse}$ & 0.85 \\
Coarse-grained frequency $s$ & $4$ \\
Fine-grained coefficient $\beta$ & 1 \\
Fine-grained threshold $I_{fine}$ & 0.98 \\
Sparse reward weight $\theta$ & 10 (if used) \\

Metaworld RL backbone & DrQv2 \\
Discount factor   & $0.98$          \\
Batch size & $256$ \\
n-step returns  & $3$ \\
Random frames in RL & $4000$ \\
Feature dimension  & $50$ \\
Task difficulty for exploration & Hard \\

Other DrQv2 hyper-parameters & See \cite{drq-v2} \\

Real-world RL backbone & HIL-SERL \\
HIL-SERL hyper-parameters & See \cite{hilserl} \\

\hline
\end{tabular}
}
\label{hyper}
\end{table}

\newpage
\subsection{Finetuning Details: Framework, Data, and Prompts }

In DEG, we employ LoRA \cite{lora} to finetune Wan2.1-I2V-14B \cite{wan} as our RL guide. We use DiffSynth-Studio (\url{https://github.com/modelscope/DiffSynth-Studio}) as our finetuning framework, following its default settings on Wan2.1-I2V-14B. The number of videos and task prompts employed in each task is provided here. 

\subsubsection{Metaworld Tasks}

\vspace{1mm}

\begin{tcolorbox}[
    colback=promptgray,
    colframe=black!20,
    arc=4pt,
    boxrule=0.5pt,
    left=6pt,
    right=6pt,
    top=6pt,
    bottom=6pt
]
\textcolor{titlegray}{
\textit{\textbf{door-close (3 videos)}}: }

\textcolor{black!70}{
A video clip of the door-close task: A red robotic arm is positioned on a table, along with a dark cabinet whose door is initially open (with its position randomly initialized and then fixed), and a light green target marker. The robotic arm moves toward the cabinet door, presses against it from the outside, and closes it. Keep the table static and the cabinet position unchanged throughout. Preserve detailed shadows on the ground. Ensure that the motion of the robotic arm and the cabinet door follows logical physics. Maintain consistent colors and appearance of objects and the scene from start to finish. After initialization, the cabinet door exhibits a slight automatic rebound, then remains stationary until contacted by the robotic arm.}
\end{tcolorbox}

\vspace{1mm}

\begin{tcolorbox}[
    colback=promptgray,
    colframe=black!20,
    arc=4pt,
    boxrule=0.5pt,
    left=6pt,
    right=6pt,
    top=6pt,
    bottom=6pt
]
\textcolor{titlegray}{
\textit{\textbf{window-close (3 videos)}}: }

\textcolor{black!70}{A video clip of the window-close task: A red robotic arm is positioned on a table alongside an open window. The robotic arm moves toward the window, presses against its white handle from the left side, and slides it to the right to close the window. Keep the table static and the window position unchanged throughout. Preserve detailed ground shadows. Ensure that the motion of the robotic arm and the window follows logical physics. Maintain consistent colors and appearance of the scene and objects from start to finish.}
\end{tcolorbox}

\vspace{1mm}

\begin{tcolorbox}[
    colback=promptgray,
    colframe=black!20,
    arc=4pt,
    boxrule=0.5pt,
    left=6pt,
    right=6pt,
    top=6pt,
    bottom=6pt
]
\textcolor{titlegray}{
\textit{\textbf{handle-press (5 videos)}}: }

\textcolor{black!70}{ A video clip of the handle-press task: A red robotic arm is positioned on a table, along with a device featuring a red handle (the device position is randomly initialized and then fixed). The robotic arm lifts up, moves above the handle, then descends and uses its gripper to press the handle downward onto the table. Keep the table static and the device position unchanged throughout. Preserve detailed ground shadows. Ensure that the motion of the robotic arm follows logical physics, and maintain consistent colors and shapes of the objects and scene from start to finish. Note that the handle does not move until the robotic arm makes contact with it.}
\end{tcolorbox}

\vspace{1mm}

\begin{tcolorbox}[
    colback=promptgray,
    colframe=black!20,
    arc=4pt,
    boxrule=0.5pt,
    left=6pt,
    right=6pt,
    top=6pt,
    bottom=6pt
]
\textcolor{titlegray}{
\textit{\textbf{button-press (3 videos)}}: }

\textcolor{black!70}{ A video clip of the button-press task: A red robotic arm is positioned on a table, along with a box featuring a red button. The robotic arm moves toward the box, accurately aligns with the red button, and presses it down. Keep the table static and the box position unchanged throughout. Preserve detailed ground shadows. Ensure that the motion of the robotic arm follows logical physics, and maintain consistent colors and shapes of the objects and scene from start to finish.}
\end{tcolorbox}

\vspace{1mm}

\begin{tcolorbox}[
    colback=promptgray,
    colframe=black!20,
    arc=4pt,
    boxrule=0.5pt,
    left=6pt,
    right=6pt,
    top=6pt,
    bottom=6pt
]
\textcolor{titlegray}{
\textit{\textbf{coffee-button (5 videos)}}: }

\textcolor{black!70}{  A video clip of the coffee-button task: A red robotic arm is positioned on a table, along with a coffee machine with a button (machine position randomly initialized and then fixed), and a coffee mug. The robotic arm reasonably adjusts its orientation until aligned with the coffee machine, then moves toward the button and presses it. Keep the table static and the coffee machine position unchanged throughout. Preserve detailed ground shadows. Ensure that the motion of the robotic arm follows logical physics, and maintain consistent colors and shapes of the objects and scene from start to finish. Note that the button does not move until the robotic arm makes contact with it.}
\end{tcolorbox}

\vspace{1mm}

\begin{tcolorbox}[
    colback=promptgray,
    colframe=black!20,
    arc=4pt,
    boxrule=0.5pt,
    left=6pt,
    right=6pt,
    top=6pt,
    bottom=6pt
]
\textcolor{titlegray}{
\textit{\textbf{drawer-open (3 videos)}}: }

\textcolor{black!70}{  A video clip of the drawer-open task: A red robotic arm is positioned on a table, along with a closed green drawer featuring a white handle. The robotic arm lifts up and moves above the white handle, then descends and uses its gripper to precisely hook the handle (with the blue inner jaw positioned inside and the white outer jaw positioned outside), and pulls the drawer open. Keep the table static. Preserve detailed ground shadows. Ensure that the motion of the robotic arm and the target object follows logical physics, and maintain consistent colors and shapes of the objects and scene from start to finish. Note that the drawer does not move until the robotic arm makes contact with it or its handle.}
\end{tcolorbox}

\vspace{1mm}

\begin{tcolorbox}[
    colback=promptgray,
    colframe=black!20,
    arc=4pt,
    boxrule=0.5pt,
    left=6pt,
    right=6pt,
    top=6pt,
    bottom=6pt
]
\textcolor{titlegray}{
\textit{\textbf{button-press-wall (3 videos)}}: }

\textcolor{black!70}{  A video clip of the button-press-wall task: A red robotic arm is positioned on a table, along with a box featuring a red button and a fixed wall. The robotic arm lifts up and closes its gripper, moves toward the box and passes over the wall, then descends to align the outer side of its gripper precisely with the red button, opens the gripper, and presses the button down. Keep the table static and the box position unchanged throughout. Preserve detailed ground shadows. Ensure that the motion of the robotic arm follows logical physics, and maintain consistent colors and shapes of the objects and scene from start to finish. Note that the red button does not move until the robotic arm makes contact with it, and the robotic arm cannot pass directly through the wall.}
\end{tcolorbox}

\vspace{1mm}

\begin{tcolorbox}[
    colback=promptgray,
    colframe=black!20,
    arc=4pt,
    boxrule=0.5pt,
    left=6pt,
    right=6pt,
    top=6pt,
    bottom=6pt
]
\textcolor{titlegray}{
\textit{\textbf{drawer-close (3 videos)}}: }

\textcolor{black!70}{  A video clip of the drawer-close task: A red robotic arm is positioned on a table, along with an open green drawer featuring a white handle. The robotic arm first lifts upward, then descends and uses its gripper to precisely press against the white handle from the outside, pushing the green drawer closed. Keep the table static. Preserve detailed ground shadows. Ensure that the motion of the robotic arm and the target object follows logical physics, and maintain consistent colors and shapes of the objects and scene from start to finish. Note that the drawer does not move until the robotic arm makes contact with it or its handle.}
\end{tcolorbox}

\vspace{1mm}

\begin{tcolorbox}[
    colback=promptgray,
    colframe=black!20,
    arc=4pt,
    boxrule=0.5pt,
    left=6pt,
    right=6pt,
    top=6pt,
    bottom=6pt
]
\textcolor{titlegray}{
\textit{\textbf{plate-slide (3 videos)}}: }

\textcolor{black!70}{  A video clip of the plate-slide task: A red robotic arm is positioned on a table, alongside a black plate initialized at a random position, and a soccer goal. The robotic arm moves downward, accurately presses on the plate, and slides it into the goal toward a red target point. Keep the table and goal static with unchanged positions. Preserve detailed shadows on the ground. Ensure the motion of the robotic arm is logical and that the colors and shapes of the objects and scene remain consistent. Note that the plate does not move until the robotic arm makes contact with it.}
\end{tcolorbox}

\vspace{1mm}

\begin{tcolorbox}[
    colback=promptgray,
    colframe=black!20,
    arc=4pt,
    boxrule=0.5pt,
    left=6pt,
    right=6pt,
    top=6pt,
    bottom=6pt
]
\textcolor{titlegray}{
\textit{\textbf{button-press-topdown (3 videos)}}: }

\textcolor{black!70}{  A video clip of the button-press-topdown task: A red robotic arm is positioned on a table, along with a box featuring a red button on its top surface. The robotic arm lifts upward, then moves over the box, aligns with the red button, and descends to press it down precisely. Keep the table static and the box position unchanged throughout. Preserve detailed ground shadows. Ensure that the motion of the robotic arm follows logical physics, and maintain consistent colors and shapes of the objects and scene from start to finish. Note that the red button does not move until the robotic arm makes contact with it.}
\end{tcolorbox}

\vspace{1mm}

\begin{tcolorbox}[
    colback=promptgray,
    colframe=black!20,
    arc=4pt,
    boxrule=0.5pt,
    left=6pt,
    right=6pt,
    top=6pt,
    bottom=6pt
]
\textcolor{titlegray}{
\textit{\textbf{hammer (3 videos)}}: }

\textcolor{black!70}{  A video clip of the hammer task: A red robotic arm is positioned on a table, along with a hammer featuring a red handle and a wooden block with a nail. The robotic arm moves toward the hammer, uses its gripper to precisely grasp the hammer’s red handle, lifts the hammer, moves it to align the hammerhead with the nail, and presses the nail into the wooden block. Keep the table static and the wooden block position unchanged throughout. Preserve detailed ground shadows. Ensure that the motion of the robotic arm and the hammer follows logical physics, and maintain consistent colors and shapes of the objects and scene from start to finish.}
\end{tcolorbox}

\vspace{1mm}

\begin{tcolorbox}[
    colback=promptgray,
    colframe=black!20,
    arc=4pt,
    boxrule=0.5pt,
    left=6pt,
    right=6pt,
    top=6pt,
    bottom=6pt
]
\textcolor{titlegray}{
\textit{\textbf{assembly (3 videos)}}: }

\textcolor{black!70}{  A video clip of the assembly task: A red robotic arm is positioned on a table, along with a circular part featuring a green handle and a fixed red cylinder. The robotic arm moves toward the part, uses its gripper to precisely grasp its green handle, lifts the part, moves it toward the red cylinder, and places the circular part onto the cylinder. Keep the table static and the red cylinder position unchanged throughout. Preserve detailed ground shadows. Ensure that the motion of the robotic arm and the part follows logical physics, and maintain consistent colors and shapes of the objects and scene from start to finish.}
\end{tcolorbox}

\vspace{1mm}

\begin{tcolorbox}[
    colback=promptgray,
    colframe=black!20,
    arc=4pt,
    boxrule=0.5pt,
    left=6pt,
    right=6pt,
    top=6pt,
    bottom=6pt
]
\textcolor{titlegray}{
\textit{\textbf{door-open (3 videos)}}: }

\textcolor{black!70}{  A video clip of the door-open task: A red robotic arm is positioned on a table, along with a black cabinet with a gray handle (cabinet position randomly initialized and then fixed), and a light green target marker. The robotic arm lifts up, moves toward the cabinet handle while gradually closing its gripper, uses the gripper to precisely press against the gray handle of the black cabinet from the inside, then pulls the cabinet door open until it reaches the position of the green target marker. Keep the table static and the cabinet position unchanged throughout. Preserve detailed ground shadows. Ensure that the motion of the robotic arm and the cabinet door follows logical physics, and maintain consistent colors and appearance of the objects and the scene from start to finish.}
\end{tcolorbox}

\vspace{1mm}

\begin{tcolorbox}[
    colback=promptgray,
    colframe=black!20,
    arc=4pt,
    boxrule=0.5pt,
    left=6pt,
    right=6pt,
    top=6pt,
    bottom=6pt
]
\textcolor{titlegray}{
\textit{\textbf{window-open (5 videos)}}: }

\textcolor{black!70}{  A video clip of the window-open task: A red robotic arm is positioned on a table, along with a closed window whose position is randomly initialized and then fixed. The robotic arm lifts up, moves toward the window, uses its gripper to press against the window's white handle from the right side, and then moves leftward to push the window open. Keep the table static and the window position unchanged throughout. Preserve detailed ground shadows. Ensure that the motion of the robotic arm and the window follows logical physics, and maintain consistent colors and appearance of the scene and objects from start to finish.}
\end{tcolorbox}

\vspace{1mm}

\begin{tcolorbox}[
    colback=promptgray,
    colframe=black!20,
    arc=4pt,
    boxrule=0.5pt,
    left=6pt,
    right=6pt,
    top=6pt,
    bottom=6pt
]
\textcolor{titlegray}{
\textit{\textbf{handle-press-side (3 videos)}}: }

\textcolor{black!70}{  A video clip of the handle-press-side task: A red robotic arm is positioned on a table, along with a dark gray device featuring a red handle (device position randomly initialized and then fixed). The robotic arm lifts up, moves above the handle, then descends and presses the handle down. Keep the table static and the device position unchanged throughout. Preserve detailed ground shadows. Ensure that the motion of the robotic arm follows logical physics, and maintain consistent colors and shapes of the objects and scene from start to finish with no pixelation or artifacts. Note that the handle does not move until the robotic arm makes contact with it.}
\end{tcolorbox}

\vspace{1mm}

\begin{tcolorbox}[
    colback=promptgray,
    colframe=black!20,
    arc=4pt,
    boxrule=0.5pt,
    left=6pt,
    right=6pt,
    top=6pt,
    bottom=6pt
]
\textcolor{titlegray}{
\textit{\textbf{faucet-close (3 videos)}}: }

\textcolor{black!70}{  A video clip of the faucet-close task: A red robotic arm is positioned on a table, along with a faucet featuring a red knob. The robotic arm moves toward the faucet, uses its gripper to precisely press against the red knob from the left side, and rotates it clockwise to close the faucet. Keep the table static and the faucet position unchanged throughout. Preserve detailed ground shadows. Ensure that the motion of the robotic arm and the target object follows logical physics, and maintain consistent colors and shapes of the objects and scene from start to finish. Note that the knob does not move until the robotic arm makes contact with it.}
\end{tcolorbox}

\vspace{1mm}

\subsubsection{Real-World Tasks}

\vspace{1mm}

\begin{tcolorbox}[
    colback=promptgray,
    colframe=black!20,
    arc=4pt,
    boxrule=0.5pt,
    left=6pt,
    right=6pt,
    top=6pt,
    bottom=6pt
]
\textcolor{titlegray}{
\textit{\textbf{real-world: pick banana (3 videos)}}: }

\textcolor{black!70}{  A video clip of the robotic arm successfully grasping and placing a banana into a plate. On a table, there are a plate, a banana initialized at a random position, and a robotic arm equipped with a gripper. The robotic arm moves horizontally above the banana, then opens its gripper, descends to precisely grasp the banana with the gripper, lifts it, moves horizontally above the plate, opens the gripper, and places the banana into the plate.}
\end{tcolorbox}

\vspace{1mm}

\begin{tcolorbox}[
    colback=promptgray,
    colframe=black!20,
    arc=4pt,
    boxrule=0.5pt,
    left=6pt,
    right=6pt,
    top=6pt,
    bottom=6pt
]
\textcolor{titlegray}{
\textit{\textbf{real-world: stack cube (3 videos)}}: }

\textcolor{black!70}{  A video clip of the robotic arm stacking a cube. On a table, there are a plate, a red cube initialized at a random position, and a wooden block. The robotic arm moves horizontally above the red cube, then opens its gripper, descends to precisely grasp the red cube with the gripper, lifts it, moves horizontally above the wooden block, opens the gripper, and places the red cube onto the wooden block.}
\end{tcolorbox}


\end{document}